%% file: main.tex
\documentclass[10pt,twocolumn,letterpaper]{article}

\usepackage[pagenumbers]{cvpr} 


\definecolor{cvprblue}{rgb}{0.21,0.49,0.74}
\usepackage[pagebackref,breaklinks,colorlinks,allcolors=cvprblue]{hyperref}

\def\paperID{*****} 
\def\confName{}
\def\confYear{2025}

\usepackage{amssymb}
\usepackage{latexsym}

\usepackage{url}
\usepackage{xcolor}

\usepackage{hyperref}
\usepackage{amsmath}
\usepackage{algorithm}
\usepackage{algpseudocode}
\usepackage{booktabs}

\definecolor{newcolor}{rgb}{.8,.349,.1}

\usepackage{amssymb}
\usepackage{latexsym}

\usepackage{framed,multirow}

\usepackage[export]{adjustbox} 

\newcommand{\bx}{\boldsymbol{x}}

\newcommand{\bt}{\boldsymbol{t}}

\newcommand{\bM}{\boldsymbol{M}}
\newcommand{\N}{\mathcal{N}}

\newcommand{\bD}{\boldsymbol{D}}

\newcommand{\bL}{\boldsymbol{L}}

\newcommand{\bs}{\boldsymbol{s}}

\newcommand{\ba}{\boldsymbol{a}}

\newcommand{\bmu}{\boldsymbol{\mu}}

\newcommand{\bphi}{\boldsymbol{\phi}}

\newcommand{\bSigma}{\boldsymbol{\Sigma}}

\newcommand{\btheta}{\boldsymbol{\theta}}

\newcommand{\bz}{\boldsymbol{z}}

\newcommand{\bv}{\boldsymbol{v}}

\newcommand{\bu}{\boldsymbol{u}}

\newcommand{\bzero}{\boldsymbol{0}}

\newcommand{\bId}{\boldsymbol{\mathbb{I}}}

\title{AtlasMorph: Learning conditional deformable templates for brain MRI}%

\author{Marianne Rakic\\
CSAIL MIT \& MGH\\
{\tt\small mrakic@mit.edu}
\and
Andrew Hoopes\\
CSAIL MIT \& MGH\\
\and
S. Mazdak Abulnaga\\
CSAIL MIT, MGH \& HMS \\
\and
Mert R. Sabuncu\\
Cornell Tech\\
\and
John V. Guttag\\
CSAIL MIT\\
\and
Adrian V. Dalca\\
CSAIL MIT, MGH \& HMS \\
\and
for the Alzheimer’s Disease Neuroimaging Initiative*
}

\begin{document}
\maketitle
\footnotetext{Data used in preparation of this article were obtained from the Alzheimer’s Disease Neuroimaging Initiative (ADNI) database (adni.loni.usc.edu). As such, the investigators within the ADNI contributed to the design and implementation of ADNI and/or provided data but did not participate in analysis or writing of this report. A complete listing of ADNI investigators can be found at: \url{http://adni.loni.usc.edu/wp-content/uploads/how\_to\_apply/ADNI\_Acknowledgement\_List.pdf}}
\begin{abstract}
Deformable templates, or atlases, are images that represent a prototypical anatomy for a population, and are often enhanced with probabilistic anatomical label maps. They are commonly used in medical image analysis for population studies and computational anatomy tasks such as registration and segmentation. Because developing a template is a computationally expensive process, relatively few templates are available. As a result, analysis is often conducted with sub-optimal templates that are not truly representative of the study population, especially when there are large variations within this population. 

We propose a machine learning framework that uses convolutional registration neural networks to efficiently learn a function that outputs templates conditioned on subject-specific attributes, such as age and sex. We also leverage segmentations, when available, to produce anatomical segmentation maps for the resulting templates. The learned network can also be used to register subject images to the templates.  We demonstrate our method on a compilation of 3D brain MRI datasets, and show that it can learn high-quality templates that are representative of populations. We find that annotated conditional templates enable better registration than their unlabeled unconditional counterparts, and outperform other templates construction methods.
\end{abstract}

\section{Introduction}
\label{sec:introduction}

Anatomical templates are representative medical images that are widely used for population studies and image analysis. 
In population studies, a template provides a common coordinate system to which the images in the study can be registered. This facilitates constructing statistical models of the population, and analyzing differences between each subject and the common template. 

Developing a template has historically involved a computationally intensive optimization process, including iterative image registration and domain-specific heuristics~\cite{allassonniere2007towards, davis2004large, de2004multi, joshi2004unbiased, ma2008bayesian, sabuncu2009image}. Because it is labor intensive and computationally expensive, there is often at most one template available per population. This is problematic for some analyses because a single template does not capture the distinctive attributes of sub-populations and can be inadequate at capturing the variability in a large dataset. 

Attempts to address this problem have typically involved identifying subpopulations, usually along a single attribute such as age, and computing separate templates for each group~\cite{sabuncu2009image,davis2010population,habas2009spatio,kuklisova2011dynamic}. This approach relies on arbitrary decisions about the attributes and thresholds used for subdividing the population. Furthermore, since each template is constructed using only images from the subpopulation, it can produce sub-optimal templates -- especially for subpopulations with a small number of images.

Templates with segmentation, or label map, are widely used by different neuroimaging software such as FreeSurfer\cite{fischl2012}, SPM~\cite{penny2011statistical}, FSL~\cite{smith2004advances} and BrainSuite~\cite{shattuck2002brainsuite}. SPM and FreeSurfer use templates involving probabilistic label maps to automatically segment new subject scans~\cite{fischl2012}. Methods often align the template to the new image and propagate the template label maps, sometimes combined with an image intensity likelihood model~\cite{puonti2016fast}. In longitudinal population studies, label maps are crucial to characterize changes in anatomy. However, template label maps are even more challenging to construct than intensity-based templates. There are often many fewer labeled images than unlabeled images for a subpopulation. Consequently, building a useful conditional atlas is more challenging when segmentation is required. 
 
\begin{figure}[t]
    \centering
    \includegraphics[width=0.45 \textwidth]{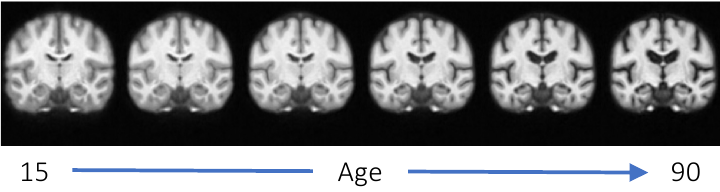}
    \caption{\textbf{Conditional Templates} developed with \textit{AtlasMorph} as a function of age. }
    \label{fig:brain_teaser}
\end{figure}

In this work, we present \textit{AtlasMorph}, a learning strategy to generate conditional templates and corresponding anatomical label maps \textit{as a function of subject attributes}. While we believe the strategy is general, in this paper we demonstrate it only on brain anatomy. Our method can leverage all the data available when constructing templates. There is no need to train a different network on a subset of the data for each combination of attributes. Figure \ref{fig:brain_teaser} shows examples of templates, generated on demand by sampling our trained age-conditioned template function. Our framework can also adjust when label maps or subject attributes are missing.

This paper builds on the methods we introduced in a preliminary conference proceeding~\cite{dalca2019learning}. This paper has the following new contributions.

\begin{itemize}
    \item We extend our probabilistic model~\cite{dalca2019learning} to characterize an anatomical label map template in addition to an intensity template. To achieve this, we adapt the learning framework to a setting where auxiliary segmentations, when available, can be used during training to build a segmentation template. Label maps are not required for test subjects during inference.
    \item We minimize the \textit{centrality} of the generated template, i.e., the weighted mean deformation between the (conditional) templates and subjects sharing the same attributes. This addresses a limitation in our previous work\cite{dalca2019learning}, where centrality was improperly modeled for \textit{conditional} templates. We introduce a new mathematical model, and a data sampling strategy based on continuous attributes, to characterize population centrality. 
    \item In a set of experiments on a varied collection of publicly available 3D brain MRI datasets, we show that the templates 1) are representative of the intended population 2) lead to higher registration quality than baseline conditional (age and sex) and unconditional templates, and 3) match better population trends than previous work~\cite{dalca2019learning}. 
\end{itemize}

\section{Related Work}
\label{sec:relatedwork}
We first discuss related work for registration, which is used both in template building and in downstream applications of templates. We then focus on previous approaches to template construction.

\subsection{Registration}
Image registration produces a dense deformation field that maps one (often called moving) image to another (fixed) image. Registration can be divided into two steps: global affine registration and deformable registration. We work with data that has already been affinely aligned, and focus on deformable transformations.

Deformable registration has been widely studied~\cite{ashburner2007fast, avants2008symmetric, bajcsy1989multiresolution, beg2005computing, dalca2016patch, glocker2008dense, thirion1998image, zimmer2019multimodal, yeo2010learning, lombaert2014spectral,zhang2017frequency,sideri2023mad}. Popular classical deformable registration methods include Demons~\cite{vercauteren2009diffeomorphic}, ANTs~\cite{avants2011reproducible}, HAMMER~\cite{shen2007image}, and ELASTIX~\cite{klein2009elastix}. These methods solve a new optimization problem for each new image pair, which makes them computationally intensive~\cite{bajcsy1989multiresolution, thirion1998image}.

In contrast, learning-based methods learn a function that for each input image pair outputs the corresponding deformation field. During inference, this provides a considerable speed up over classical methods~\cite{balakrishnan2018unsupervised, balakrishnan2019voxelmorph, cao2017deformable, dalca2019unsupervised, krebs2017robust, rohe2017svf, sokooti2017nonrigid, vos2017end, yang2017quicksilver, lee2019image, evan2022keymorph, krebs2019learning, hoffmann2021synthmorph,hoopes2021hypermorph, mok2021conditional}. Some learning-based methods are supervised, where ``ground truth'' deformation fields are precomputed or synthesized~\cite{dosovitskiy2015flownet, cao2018deformable, yang2017quicksilver, lee2019image}. Unsupervised methods learn to predict a dense deformation field without the corresponding ground truth~\cite{balakrishnan2018unsupervised, fan2019adversarial, de2019deep, krebs2019learning, qin2019unsupervised, mansilla2020learning}. Instead, during training they usually employ loss terms to encourage 1) matching between the warped source and the target and 2) regularity of the deformation field. 

Some methods leverage image segmentations as auxiliary labels~\cite{balakrishnan2019voxelmorph,dalca2019unsupervised, hu2018weakly, xu2019deepatlas, hoffmann2021synthmorph}. These labels can be used in an additional loss term at training time to encourage the anatomical structures between the fixed and moving image to match. The auxiliary labels are not used at inference time. Image segmentations or label maps are often hard to obtain since they require manual annotations by experts. But when available at training time, they can improve registration performance.

Using deformation fields to map one image to another is an ill-posed problem. Different deformation fields can yield the same moved image. An unconstrained mapping of voxels from one image to the other carries little information about the underlying similarity of the anatomy. Spatial regularization strategies have been widely studied to address these issues. Some restrict the set of valid deformation fields, for example, by using B-splines~\cite{rueckert1999nonrigid}. Others add a term in the loss function, encouraging small spatial gradients on the displacements~\cite{balakrishnan2019voxelmorph,mok2022unsupervised}, smooth velocity fields~\cite{krebs2018unsupervised,hoopes2021hypermorph}, or small bending energy~\cite{de2019deep, staring2007rigidity}.

In medical imaging, it is often valuable to preserve topology and ensure that the deformation fields are anatomically plausible. One way to accomplish this is to work with diffeomorphic deformation fields~\cite{avants2008symmetric, ashburner2007fast, beg2005computing, cao2005large, ceritoglu2009multi, hernandez2009registration, joshi2000landmark,miller2005increasing, oishi2009atlas, vercauteren2009diffeomorphic, zhang2017frequency}. Diffeomorphisms are differentiable and invertible deformation fields, which are therefore topology preserving. In this framework, one option is to parametrize the deformation field using a stationary velocity field~\cite{arsigny2006log, ashburner2007fast, vercauteren2008symmetric, bossa2010tensor}. To approximate stationary fields,  we use Scaling and Squaring~\cite{ashburner2007fast,vercauteren2009diffeomorphic,hernandez2009registration,arsigny2006log,arsigny2006logpolyaffine,arsigny2007geometric}.

In our work, we build on learning based registration frameworks to learn conditional templates. We employ label maps when available, and use diffeomorphic deformation fields to ensure anatomical fidelity.

\subsection{Template Construction}
Templates are often used in population studies, where they serve as a central reference when comparing images from the subject population. To compute a template, iterative methods start with an initial template, such as the voxel-wise average over the population or a randomly selected image from the population. Then, through an iterative process, they register the images in the population to the intermediate template, compute a new average template from the aligned images, and iterate until convergence~\cite{allassonniere2007towards,davis2004large,joshi2004unbiased, ma2008bayesian,sabuncu2009image,lombaert2012human,guimond1998automatic}. This process is very computationally expensive. 

Often, a single template is not sufficient to characterize an entire population. The template might be too different from some subjects to get a meaningful comparison. To address this problem, some studies employ different templates for population subgroups such as infants and the elderly~\cite{prastawa2005automatic,thompson2001cortical}. However, such templates can be less representative since they are often built from insufficient data. In addition, dividing the population into subgroups can be challenging for continuous attributes such as age. Some methods use clustering~\cite{sabuncu2009image} or arbitrary thresholds, but this leads to detrimental discontinuities in downstream studies because subjects might be matched with a relatively distant template. 

Some methods~\cite{davis2010population, habas2009spatio, kuklisova2011dynamic} tackle scenarios where the type of variability is explicitly modeled, such as spatiotemporal brain templates. They use a combination of domain knowledge, manual anatomical segmentation, and extensive computational resources to produce the templates, using classical methods.

In our preliminary work~\cite{dalca2019learning}, we showed that it was possible to learn templates using neural networks, and optionally to condition them on available attributes~\cite{dalca2019learning}, leading to faster template construction and conditioning on desired attributes. Building on this, several strategies have improved different aspects~\cite{Dey_2021_ICCV, sinclair2022atlas, he2020unsupervised,  he2021learning, ding2022aladdin}, including template realism. Some learn an unconditional template made of label maps only~\cite{sinclair2022atlas}. Recent work uses Generative Adversarial Networks templates that appear more like an acquired image~\cite{Dey_2021_ICCV}. This work uses a discriminator to distinguish between the real subject scan and the deformed template, but does not produce label maps. 

One line of work trains a large network containing different specialized subnetworks: for instance, one that specializes in segmentation, the other in registration~\cite{lee2019image,  xu2019deepatlas}. The intent is to combine the advantages of each subnetwork to obtain better overall performances for both registration and segmentation. For example, the \textit{image-and-spatial-transformer} network (ISTN)~\cite{lee2019image} combines an image transformer network and a spatial transformer network. The image transformer network learns an intermediate representation of the input image, which is then passed to the spatial transformer network, which performs registration. ISTNs were extended~\cite{sinclair2022atlas} to jointly train a segmentation, registration, and an unconditional template building network. Their resulting template includes both a voxel representation and the corresponding label map, for cardiac tomography. The work focuses on building a single template for the population. In contrast, we focus on expanding \textit{conditional} template construction, which enables much richer template functions.

\begin{figure*}[t]
    \centering
    \includegraphics[width=0.9\textwidth]{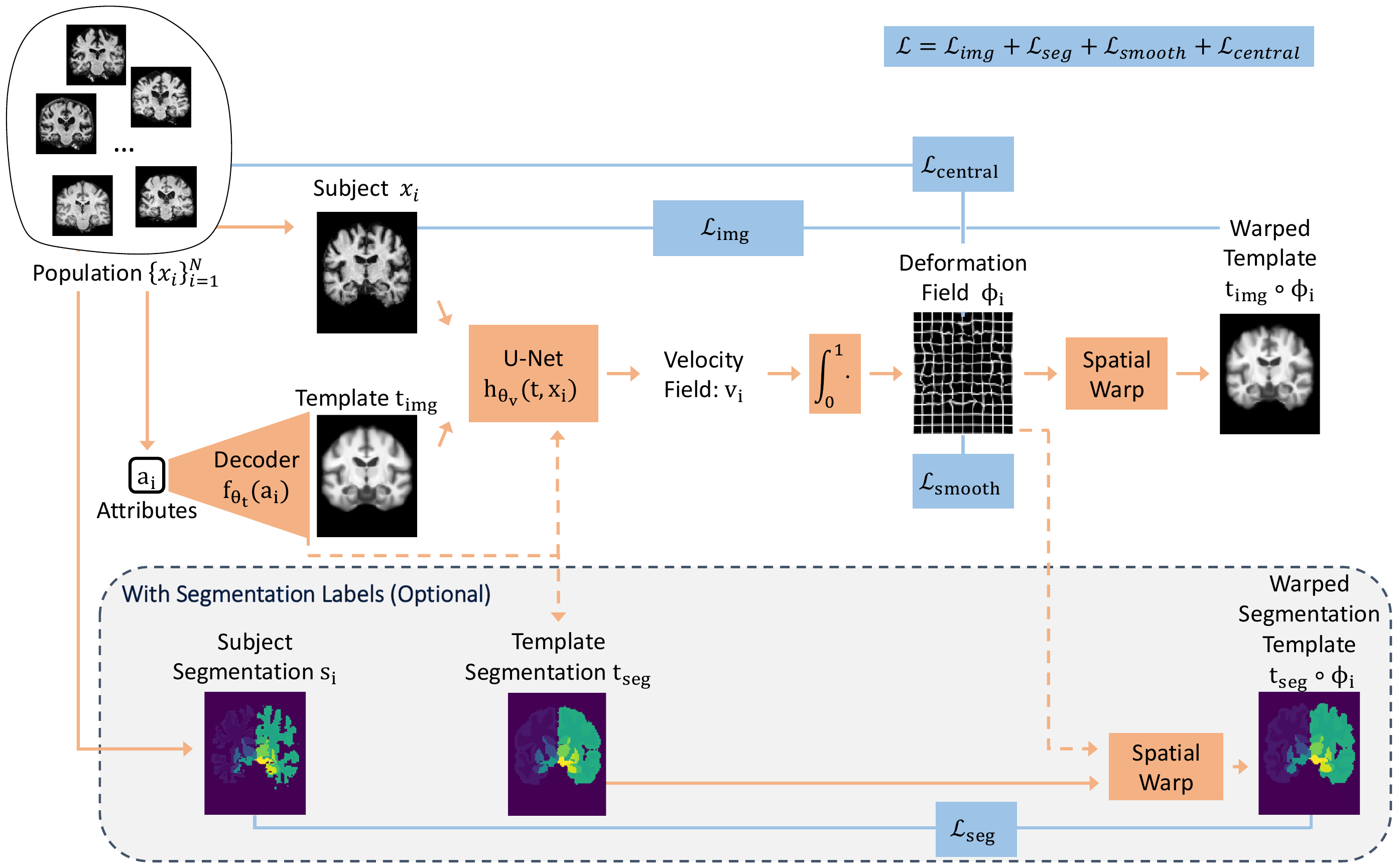}
    \caption{\textbf{\textit{AtlasMorph} Architecture.} We employ two main networks, a decoder which creates templates (3D volumes and corresponding segmentation maps) given an attribute, and a registration network that aligns subjects to templates. We learn these jointly using a loss that is a function of the subject scan, the warped template and the deformation field. The decoder and the UNet contain learnable weights. We estimate both a voxel template and the corresponding label maps, shown with the shaded \textit{optional} box.}
    \label{fig:archicture:total}
\end{figure*}

\section{Method}
We describe a generative probabilistic model of anatomical conditional templates, with corresponding probabilistic segmentation maps. 
We design a new sampling strategy to properly approximate centrality of conditional templates, and derive a new loss function for conditional templates. We show that with this framework the unconditional and unlabelled templates are special cases of conditional template functions. Finally, we describe the neural network architecture and parameters.

\subsection{Probabilistic model}
\label{subsubsec:proba_modelcond}
Let $\mathcal{Z}=\{\bz_i\}_{i=1}^N$ be a dataset containing sample image-segmentation pairs~$\bz_i=(\bx_i, \bs_i)$. We let~$\mathcal{A} = \{\ba_i\}$ be the set of attributes corresponding to subject images~$\mathcal{X}$, representing non-image subject-specific information available in the data. We model a \textit{conditional} template as $\bt = f_{\btheta_{\bt}}(\ba)$, a function of the attributes $\ba$ parametrized by~$\btheta_t$. The template contains both an image representation and a matching label map representation: $\bt = (\bt_{img}, \bt_{seg})$.

We model each sample image-segmentation pair as a noisy observation of a spatial deformation~$\bphi_{v_i}$ of the conditional template~$\bt$. For example, for the image component, $\bx_i = \bt_{img} \circ \bphi_{v_i} + \epsilon_i$, where $\circ$ is a spatial warp operation, the deformation field $\bphi_{v_i}$ is parametrized by the vector field $v_i$ and $\epsilon_i$ is Gaussian noise. We model the function $ h_{\theta_v}(\bt, \bx_i)=v_i$, parametrized by $\theta_v$ that estimates the velocity field between an image and template. 

We maximize the probability of the data to estimate the model parameters~$\btheta = (\btheta_{\bt}, \btheta_v)$, 
\begin{align}
\hat{\btheta~} &= \arg {\max_{\btheta}} {\log p_{\btheta}}(\mathcal{V} | \mathcal{Z}; \mathcal{A}) \nonumber \\
	&= \arg {\max_{\btheta}} {\log p_{\btheta}}(\mathcal{Z} | \mathcal{V}; \mathcal{A}) + \log p_{\btheta}(\mathcal{V|\mathcal{A}}).
\label{eq:logpostcond}
\end{align}
where the first term represents the data likelihood, and the second term is a prior over the velocity fields, and $\mathcal{V} = \{v_i\}$.

\subsubsection{Deformation fields} We employ diffeomorphic deformation fields. We treat $v$ as a stationary velocity field~\cite{ashburner2007fast, balakrishnan2018unsupervised, hernandez2009registration, krebs2018unsupervised, modat2012parametric} and define deformation field $\bphi_v$ as:
 \begin{align}
     \frac{\partial \bphi_v^{(t)}}{\partial t} = \bv (\bphi_v^{(t)}),
 \end{align}
 with time variable $t$.
 The deformation field, $\bphi=\int_0^1 \bv \, dt$, is then obtained by integrating the velocity field, $\bv$, using Scaling and Squaring~\cite{ashburner2007fast,arsigny2006logpolyaffine,arsigny2007geometric,arsigny2006log,vercauteren2009diffeomorphic,hernandez2009registration,dalca2019unsupervised}.

We decompose the deformation field as: $\bphi_v = Id + \bu_v$, where $\bu_v$ is the spatial displacement and $Id$ is the identity transform. Building on~\cite{dalca2019unsupervised}, we model the prior over the velocity field as: 
\begin{align}
p(\mathcal{V}| \ba)  \propto  \exp\{- \lambda_c \| \bar{\bu}_v(\ba)\|^2\}  \prod_i \N(\bu_{v_i}; \bzero, \bSigma_{u_{v_i}}),  
\label{eq:prior}
\end{align}
where~$\mathcal{N}(\cdot;\bmu_v,\bSigma)$ is the multivariate normal distribution with mean~$\bmu$ and covariance~$\bSigma$, and~\mbox{$\bar{\bu}_v(\ba)$} describes the average displacement field between the template and subjects with attribute values $\ba$. We let~$\bSigma^{-1}=\bL$, where~$\bL = \lambda_d \bD - \lambda_a \bM$ is a relaxation of the Laplacian of a neighborhood graph defined on the voxel grid, with~$\bM$ the voxel neighborhood adjacency matrix, and~$\bD$ the graph degree matrix~\cite{dalca2019unsupervised}. The hyperparameters $\lambda_{c}$, and~$\lambda_a$ represent the relative weight for the centrality of the template, and the smoothness of the deformation fields, respectively.

Let $\nabla \bu_{v_i}$ be the spatial gradient of the displacement field. The log prior over the velocity field is then: 
\begin{align}
\log p(\mathcal{V}| \ba)  =  -\lambda_{c} \| \bar{\bu}_v(\ba) \|^2   - \sum_i \frac{\lambda_a}{2} \| \nabla \bu_{v_i} \|^2 + \text{const}.
\end{align}
The first term
assesses the centrality or bias of the template by measuring the \textit{average} displacement. The second term  evaluates the smoothness of deformations from the template. 

\subsubsection{Data Likelihood} 
For images, we use an additive Gaussian model:
\begin{equation}
p(\bx_i | \bv_i ; \ba_i) = \N(\bx_i ; {f_{{\btheta{}_{\bt}}}}(\ba_i) \circ  \bphi_{v_i}, \sigma^2 \bId ).
\end{equation}
where~$\sigma^2$ accounts for additive image noise. 

Inspired by Dice score~\cite{dice1945measures}, we model the probability of the label maps as: 
\begin{equation}
p(\bs_i | \bv_i ; \ba_i) \propto \exp\left\{\frac{2 \lvert \bs_i  \cap f_{\btheta_{\bt}}(\ba_i)[seg] \circ \bphi_{v_i} \rvert}{\lvert\bs_i\rvert+\lvert f_{\btheta_{\bt}}(\ba_i)[seg]\circ \bphi_{v_i}\rvert}\right\},
\end{equation}
where $f_{\btheta_{\bt}}(\ba_i)[seg]$ is the label map component of $f_{\btheta_{\bt}}(\ba_i)$.

\subsection{Loss}
To find the optimal parameters $\btheta = ({\btheta_{\bt}}, \btheta_v)$, we minimize the negative maximum likelihood~\ref{eq:logpostcond}:
\begin{align}
    \mathcal{L}(&{\btheta}; \mathcal{V}, \mathcal{Z}, \mathcal{A}) = -\sum_i \log p_{\theta}(\bv_i , {\bz_i}; \ba_i).
\end{align}
Focusing on individual terms, we have:
\begin{align}
	\mathcal{L}(&{\btheta}; \bv_i, \bz_i, \ba_i)= -\log p_{\theta}(\bv_i , {\bz_i}; \ba_i)\\
	&= -\log p_{\theta}(\bz_i | \bv_i; \ba_i) - \log p_{\theta}(\bv_i) \nonumber \\ \nonumber
	&=  \frac{\lambda_{img}}{2 \sigma^2} \| \bx_i - f_{\bt}(\ba_i) \circ \bphi_{v_i} \|^2  &
 (\mathcal{L}_{img})\\\nonumber
 &\quad\quad - \lambda_{seg} \frac{2 \lvert \bs_i  \cap f_{\btheta_{\bt}}(\ba_i)[seg] \circ \bphi_{v_i} \rvert}{\lvert\bs_i\rvert+\lvert f_{\btheta_{\bt}}(\ba_i)[seg]\circ \bphi_{v_i}\rvert} &
 (\mathcal{L}_{seg})\\\nonumber
  &\quad\quad +\lambda_c \| \bar{\bu}_v (\ba_i)\|^2 &
 (\mathcal{L}_{central})\nonumber \\
	&\quad\quad + \sum_j \frac{\lambda_a}{2} \| \nabla \bu_{v, j} \|^2  + \text{const} &
 (\mathcal{L}_{smooth})\nonumber
\end{align}
where $\lambda_{img}$, $\lambda_{seg}$ are hyperparameters controlling the relative importance of each term in the loss function. This expression can be generalized to other $\mathcal{L}_{img}$ and $\mathcal{L}_{seg}$ terms such as negative cross-corelation and mutual information, or categorical cross-entropy, respectively. 
\subsection{Centrality Loss Term}
\label{methods:centrality}
In the preliminary work~\cite{dalca2019learning}, we used a global centrality $\bar{\bu}(\ba_i) \approx \bar{\bu} = \sum_{k=1}^K \bu_{k}$ to approximate the centrality at each attribute. However, the average deformation field can vary dramatically by attribute value, leading to this approximation constraining resulting templates. To address this, we formulate the centrality measure to only compare deformation fields from the same conditional template. We approximate this in practice using kernel density estimation, with a Gaussian kernel. This takes into account the distribution of scans available around a given attribute. 

When the attribute~$\ba$ is categorical, we define $$\bar{\bu}_v(\ba) = \sum_{i\in\mathcal{S}(a)}^K \bu_{v, i},$$
for $\mathcal{S}(\ba)$=\{$j |$ subject $j$ has attribute value $\ba$\}. 

When $\ba$ is continuous, we approximate conditional centrality by averaging the displacement fields of subjects with similar attributes using kernel density estimation around $\ba$. For each subject $i$ with attribute value $\ba_i$, we first compute the data density $Q_i$ around $\ba_i$:
\begin{equation}
    Q_i(a) = \sum_{j \neq i } \exp \left(-\frac{1}{\sigma_d} (a_i -a_j)^2 \right),
\end{equation}
where $\sigma_d$ is a hyperparameter. Next, for any attribute value $a$, we define the weight $w_i(a)$ of any subject $i$:
\begin{equation}
    w_i(\ba) = \frac{1}{Q_i}\exp\left\{-\frac{1}{\sigma}(\ba-x_i)^2\right\},
\end{equation}
where $\sigma$ is a hyperparameter. Finally, we let the conditional centrality be the weighted average: 
\begin{equation}
    \bar{u}(\ba) = \frac{1}{\sum_iw_i(\ba)}\sum_i w_i(\ba) \,\bu_i.
\end{equation}
In our implementation, we approximate the centrality loss term by only sampling B subjects within a categorical attribute at each iteration. For continuous attributes, we sample subjects with probability $\frac{w_i(\ba)}{\sum_i w_i(\ba)}$.

This centrality term is only used in the loss function to encourage centrality, and no deformation field directly results from it.

\subsection{Neural Network Model}
Overall, \textit{AtlasMorph} takes the form $g_{\btheta}(\bx_i, \ba_i) = (v_i, \bt)$, which we factorize as $\bt=f_{\btheta_t}(\ba_i)$,   $v_i = h_{\btheta_v}(\bt, \bx_i)$. We model $f$ and $h$ as neural networks. The first network, $f_{{\btheta}_{\bt}}(\ba)$, generates the template $\bt$, both the intensity voxels and label probability maps, using parameters $\theta_{\bt}$. The second network, $h_{\theta_v}(\bt, \bx_i)$ estimates the velocity field $v$ that aligns generated template $\bt$ to subject $i$. For this sub-network, we use a UNet-like architecture similar to VoxelMorph \cite{dalca2019unsupervised}. Rather than getting two images as input, as VoxelMorph does, our network takes in one image and has to learn the other (the template) by aligning it to all subjects in the population with corresponding attributes. We use one-hot-encoding to model the labels for each scan. The deformation fields produced are diffeomorphic up to the numerical errors resulting from numerical approximations when using Scaling and Squaring. The term $\mathcal{L}_{smooth}$ encourages smooth deformation fields. The loss term $\mathcal{L}_{central}$ encourages the sum of the deformation fields to be close to 0, which we interpret as a measure of template centrality. The \textit{AtlasMorph} architecture is pictured in Figure \ref{fig:archicture:total}.

\subsection{Inference} 
During inference, we obtain the template from $f_{\btheta_{\bt}}(\cdot)$, using a single forward pass of the network, and providing the attributes $\ba$, if available. A forward pass through the registration sub-network $h_{\btheta_v}(\cdot, \cdot)$ yields the deformation field $\phi_{v_i}$ between the template $\bt$ and an image sample $\bx_i$, by integrating the resulting velocity field $v_i$. 
Since we use diffeomorphic deformation fields, we obtain the inverse deformation field $\phi_{v_i}^{-1}$ (from the sample image to the template), by negating the velocity field and then integrating it: $\phi_{v_i}^{-1}=\phi_{-v_i}$.

\subsection{Model Variants}

If no conditioning attributes are available, we model $f_{\btheta_{\bt}}(\cdot)$ as a set of learnable parameters representing the voxel intensities directly, with each parameter representing each voxel intensity in the learned template image.

When label maps are unavailable during training, our network learns the intensity template, without the corresponding segmentation template. The segmentation loss term, $\mathcal{L}_{seg}$ is ignored at training by setting $\lambda_{seg}$ to 0. We then use these inverse deformation fields to obtain the template label maps.

\subsection{Implementation}
We use the same implementation of Scaling and Squaring as~\cite{dalca2019unsupervised}. We implement the registration network $h_{\btheta_v}(\bt, \bx)$ as a 3D UNet with the following design: 
\begin{itemize}
    \item Encoder-Decoder channels with 4 layers each. The encoder and decoder have [16, 32, 32, 32] and [32, 32, 32, 32, 32, 16, 16] features per layer respectively.
    \item Convolutional layere have kernels of size 3 and ReLU activations.
\end{itemize}
The decoder, $f_{\theta_{\bt}}(\ba)$, has the following components: \begin{itemize}
    \item A dense layer, with the attribute as input, and the number of voxels in the template divided by 64 as output.
    \item Three layers alternating convolutional layers, max-pooling and upsampling by 2. The convolutional layers have kernel size 3 and ReLU activation.
\end{itemize}

\noindent \textbf{Training.} We use the Adam optimizer with a learning rate of 0.0001 and a batch size $B$ of 3, the maximum we could fit in GPU memory of an A100. We train the model for 4000 epochs until convergence, which corresponds to a week. We ran inference on an A100 with 40GB of memory.
We train by using categorical crossentropy segmentation loss until convergence.
After a grid search, we use the following loss hyperparameters: $\sigma=2$, $\sigma_d=1$, $\lambda_{img}=20$, $\lambda_{seg}=0.2$, $\lambda_{a}=1$, $\lambda_{c}=0.1$. In our implementation, we approximate the centrality loss term by only sampling $B$ subjects within a categorical attribute, and sampling subjects with probability $\frac{w_i(\ba)}{\sum_i w_i(\ba)}$ for continuous attributes. Since the batch samples are randomly selected at each training iteration, all the samples in the training set are eventually taken into account in the loss.

%
\section{Experiments}
Using a collection of 3D brain MRI datasets, we assess the learned templates in terms of registration accuracy and field regularity. We also analyze the learned anatomical structures with respect to the population trend, and that our updated centrality loss captures population characteristics better than the previous preliminary model~\cite{dalca2019learning}.

\subsection{Data} 
We use 10,195 T1-weighted 3D brain MRI scans from the publicly available datasets ADNI~\cite{mueller2005ways}, OASIS~\cite{marcus2007open} as prepared in~\cite{hoopes2021hypermorph}, ABIDE~\cite{di2014autism}, and IXI~\cite{ixidata}.

We perform standard neuroimaging pre-processing steps. This includes resampling to~$1$mm isotropic voxels, affine alignment, normalization, skull stripping and anatomical segmentations using FreeSurfer~\cite{fischl2012}. We crop the final volumes to $160\times 192 \times 224$ voxels, used for all our experiments. The dataset is split by subject into 8188 training volumes, 1007 validation volumes, and 1000 test volumes belonging to 2947, 368, and 370 subjects in the training, validation, and test set, respectively. In the training set, we have 6637 scans from ADNI, 751 from ABIDE, 450 from IXI, and 332 from OASIS. Final evaluation is done on the held-out test set of 1000 subjects for all baselines and \textit{AtlasMorph} variants. 

We conduct our analysis on 17 anatomical structures based on the FreeSurfer protocol. For anatomical regions defined in both brain hemispheres, such as left and right ventricles, we consider it as a single structure when reporting results. 

\begin{figure}
    \centering
    \includegraphics[width=.23\textwidth]{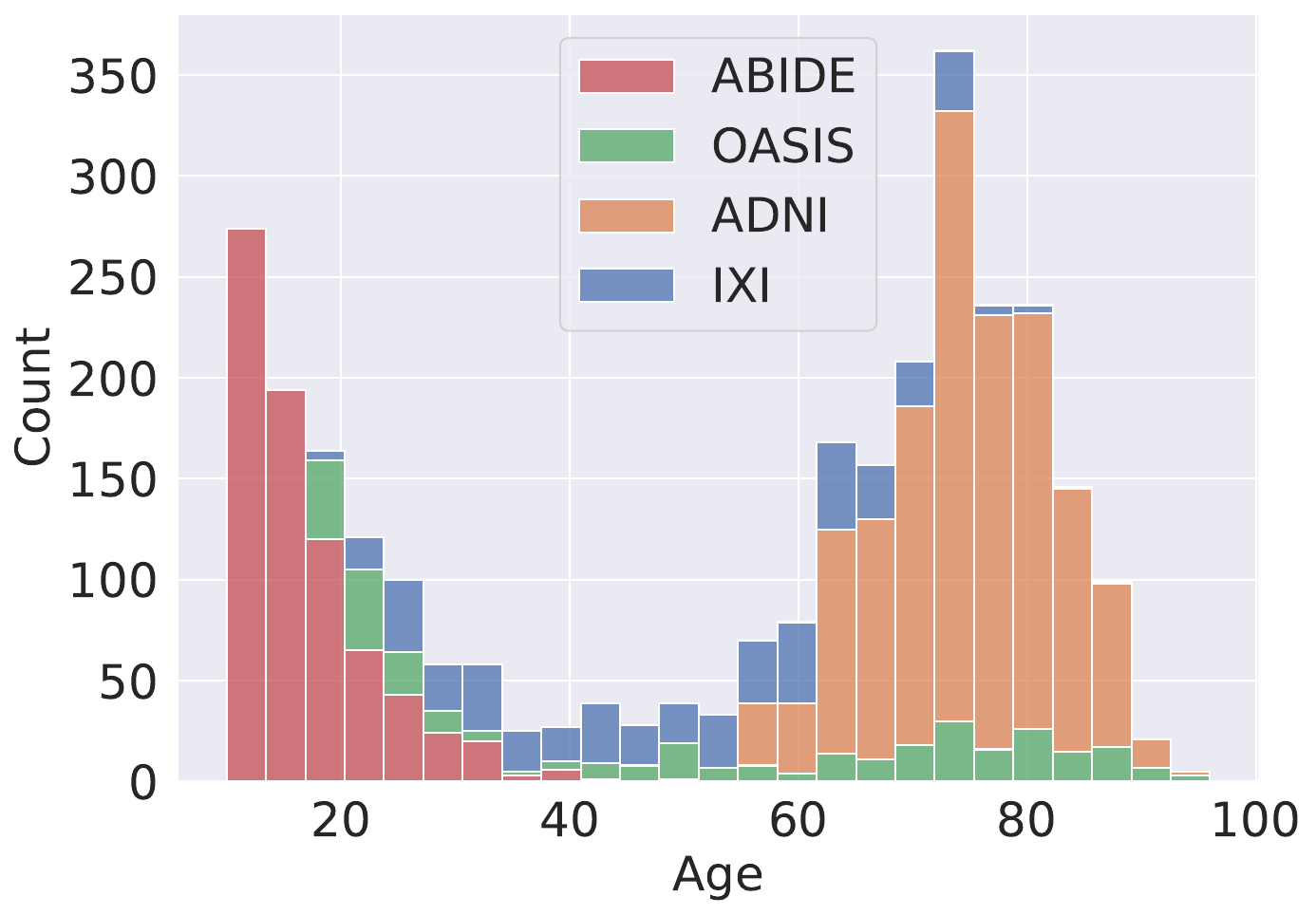}
    \hfill
    \includegraphics[width=.23\textwidth]{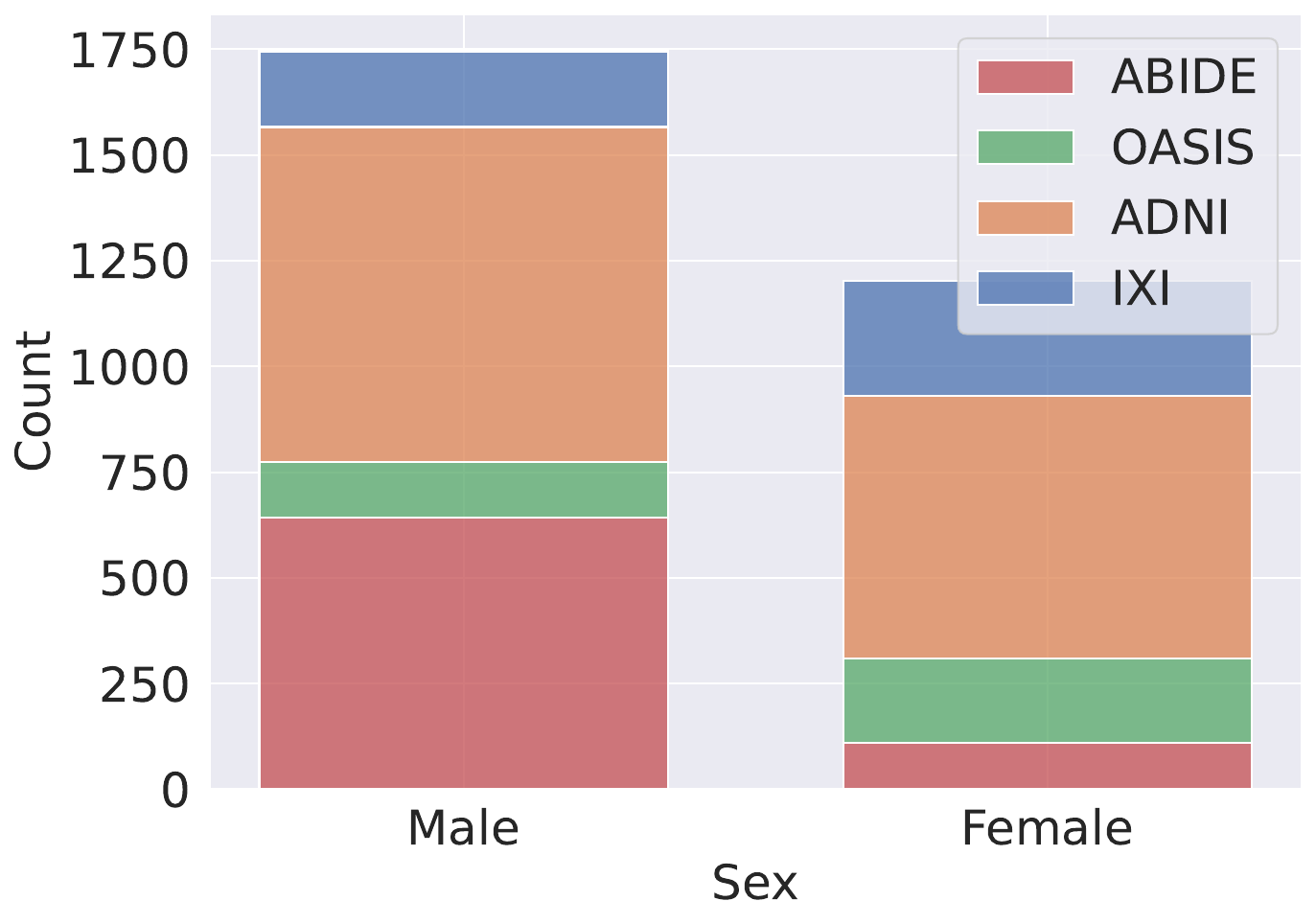}
    \caption{\textbf{Statistics for subjects in the training set.} Left: Distribution of subject age. Right: Distribution of subject sex.}
    \label{fig:data:attribute_stats}
\end{figure}

 We analyze age and sex subject-specific attributes, as defined in each of the datasets~\cite{mueller2005ways,marcus2007open,di2014autism,ixidata}. The data covers ages from 12 to 90 years old for both male and female subjects. Figure \ref{fig:data:attribute_stats} shows the distribution for both attributes.

\subsection{Set up}
\noindent\textbf{Metrics:} There is no single metric that fully captures (or  is widely used to compare) anatomical templates. We compare templates using registration accuracy, field regularity, atlas centrality, atlas sharpness, and the characterization of population trends.

\noindent\textbf{Baselines.} We compare \textit{AtlasMorph} to seven methods for template-building. There are two phases: acquiring the template or templates, and producing the deformation fields between the template and corresponding subjects.

\begin{figure}[h!]
    \centering
    \includegraphics[width=0.45\textwidth]{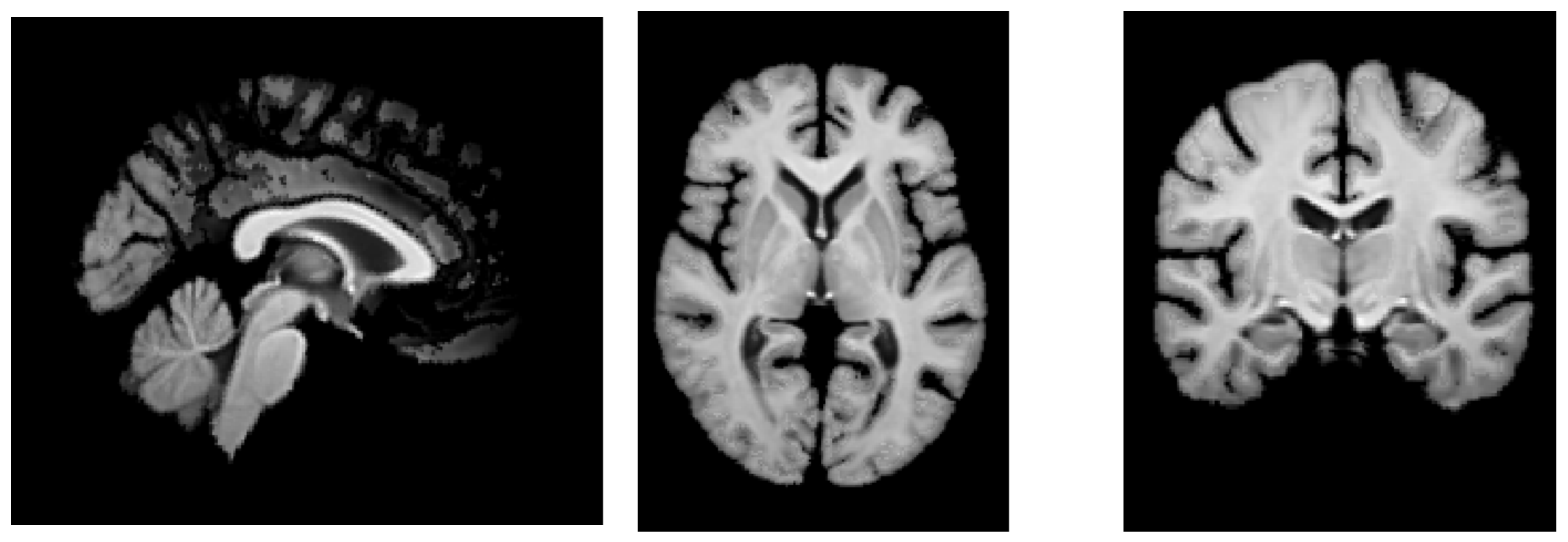}
    \caption{\textbf{Iterative template used in \textit{VxM-Single-no-Seg}}, build using classical template building techniques. }
    \label{fig:baseline:3D_old_template}
\end{figure}

\begin{itemize}

    \item \textit{ANTs-Single-no-Seg}: We use ANTs~\cite{avants2011reproducible} to compute a template from the training data. We synthesize the template from 100 training subjects randomly sampled. Details of the ANTs parameters used are described in Supplementary Section~\ref{s:unconditional-ants}. We then use ANTs to register the synthesized template to the test subjects.
    
    \item \textit{ANTs-Frag-no-Seg}: We build six different templates by dividing the population using three age groups and male versus female. Details are described in Supplementary Section~\ref{s:conditional-ants}. We use ANTs to register the produced templates to the corresponding images, matching the subjects to their corresponding templates.
    
    \item \textit{ANTs+VxM-Single-no-Seg}: We use ANTs~\cite{avants2011reproducible} to compute a template from our training data. We synthesize the template from 100 training subjects randomly sampled. Details of the ANTs parameters used are described in Supplementary Section~\ref{s:unconditional-ants}.
    
    To produce the corresponding deformation fields, we train a \textit{VoxelMorph} model to register the fixed template to all the training images available, using the same UNet model size as used in \textit{AtlasMorph}. Finally, we obtain the label maps for each template by registering the 100 subjects used for atlas construction to the resulting atlas, propagating the corresponding labels and averaging the propagated labels.

    \item \textit{ANTs+VxM-Frag-no-Seg}: We build six different templates by dividing the population using three age groups and male versus female. Details are described in Supplementary Section~\ref{s:conditional-ants}. To produce the corresponding deformation fields, we train a VoxelMorph network similarly to the \textit{ANTs+VxM-Single-no-Seg} baseline and adjust the template based on the corresponding attributes.
    
    \item \textit{Aladdin-Single}: Aladdin~\cite{ding2022aladdin} is a recent learning based method that learns to jointly register an image and produce a template for the corresponding image. We train an Aladdin model with the same capacity on \textit{AtlasMorph} and on the same training data.
    \item \textit{Aladdin-Frag}: To produce templates that vary with age and sex for Aladdin, we bin the population similarly to \textit{Frag-ANTs} baselines.

    \item \textit{ANTs+VxM-Single-no-Seg, B40.}: We use the template~\cite{sridharan2013quantification} used  in \textit{VoxelMorph}~\cite{balakrishnan2019voxelmorph,dalca2019learning}, shown in Figure \ref{fig:baseline:3D_old_template}. This template was built using ANTs on 39 subjects from Buckner40 data~\cite{admiraal2005fully,johnson1998preclinical,rj2000use}. To produce the corresponding deformation fields, we train a VoxelMorph networks similarly to the \textit{ANTs+VxM-Single-no-Seg} baseline.
\end{itemize}

\begin{figure}[t]
    \centering
    \includegraphics[width=.45\textwidth]{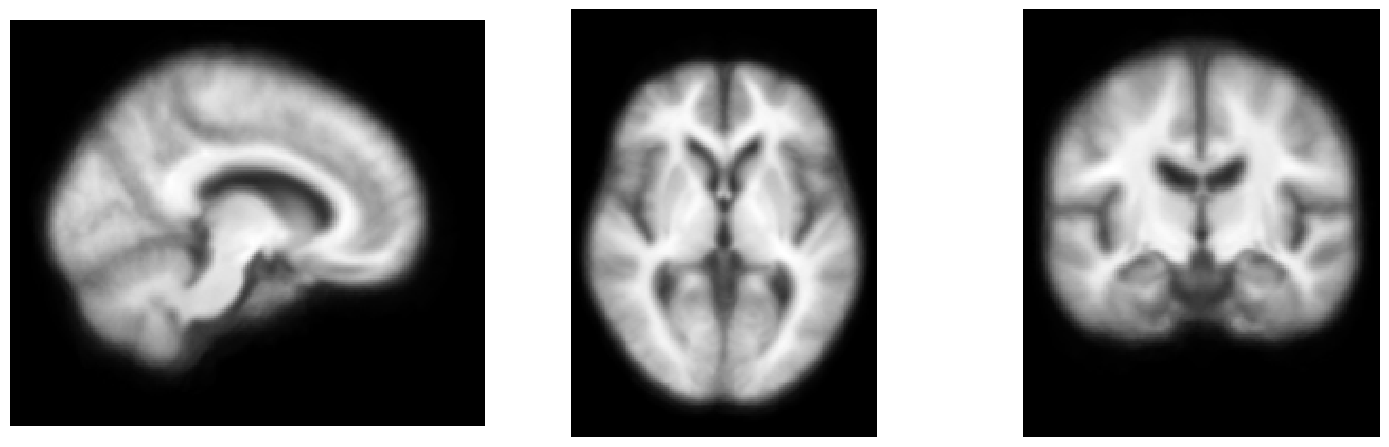}
    \includegraphics[width=.45\textwidth]{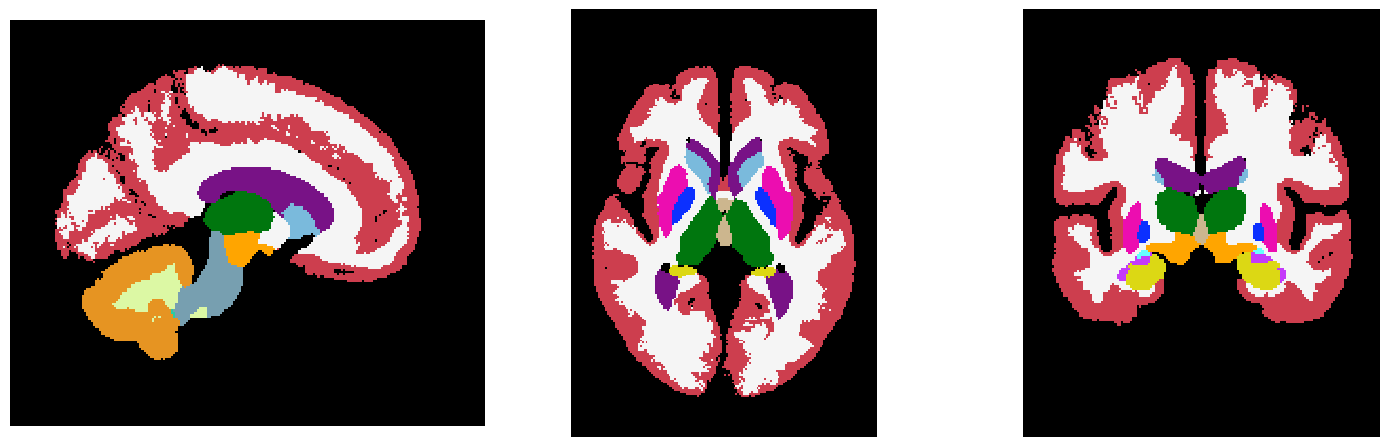}
    \caption{\textbf{Template construction by averaging the first 100 subjects in our training data.} Top: Intensity template obtained by averaging. Bottom: Corresponding label map to the average intensity template. The templates constructed by averaging is notably less sharp than the \textit{AtlasMorph} templates.}
    \label{fig:exp:3Daverage}
\end{figure}

\begin{figure*}
    \centering
    \includegraphics[width=0.87\textwidth]{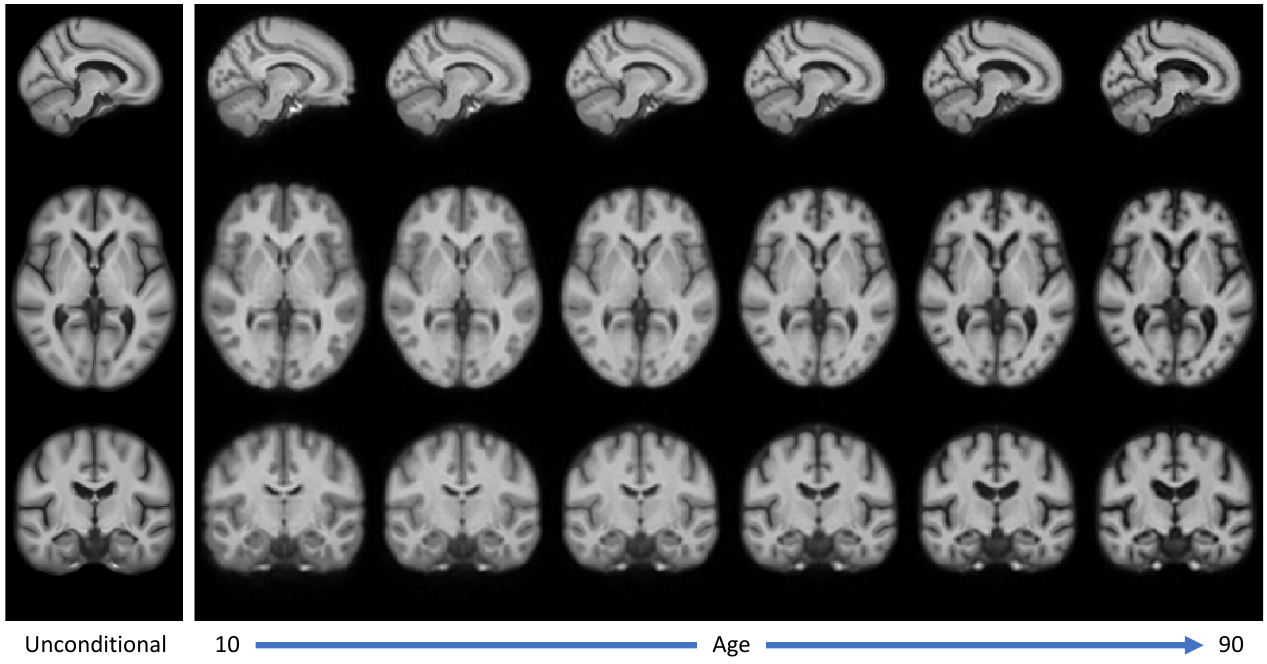}
    \caption{\textbf{Intensity Templates obtained with \textit{AtlasMorph}.} Left: Unconditional Intensity Template. Right: Conditional Intensity Templates sampled from our learned template function from age 10 to 90, left to right. The conditional templates capture well-known signs of age-associated atrophy as analyzed in Figure \ref{fig:3D_vent_vol_kernel}.}
    \label{fig:3D_cond_img}
\end{figure*}
\begin{figure*}
    \centering
    \includegraphics[width=0.87\textwidth]{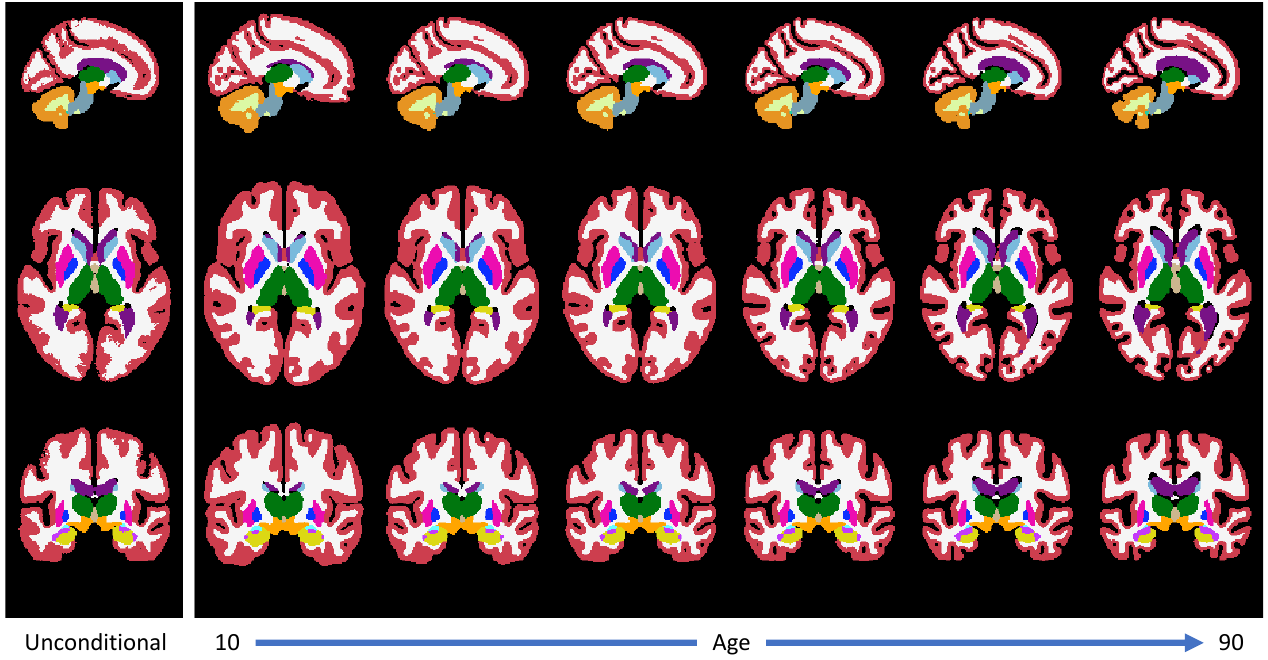}
    \caption{\textbf{Label Maps Templates obtained with \textit{AtlasMorph}.} Left: Unconditional Label Maps Template. Right: Conditional Label Maps. Templates sampled from our learned template function from age 10 to 90, left to right. The label maps visualizations are obtained by assigning to each class the label with the highest probability. The conditional templates capture well-known signs of age-associated atrophy.}
    \label{fig:3D_cond_seg}
\end{figure*}

To better understand our dataset variability and estimate the visual difference/improvement for any downstream solution, we also compute and visually assess the template produced by averaging the first 100 subjects in our dataset without registration, but omit it from quantitative analysis as the quality is too low.\\

\noindent\textbf{Ablations.} We assess the importance of each component of our framework using the following \textit{AtlasMorph} variants: 
\begin{itemize}
    \item \textit{AtlasMorph} (AM) - the full proposed model using subject age and sex attributes. 
    \item \textit{AtlasMorph-no-Seg} learns a template function, without segmentation labels. For evaluation, we compute a label map  \textit{after training}, by registering each MRI in our training set to the template, propagating the corresponding labels from the brain MRI to the template, and averaging the propagated labels.
    \item \textit{AtlasMorph-Uncond} learns an unconditional template with segmentation labels. The template label map is computed at the same time as the template image by a forward pass of the template subnetwork.
    \item \textit{AtlasMorph-Uncond-no-Seg} learns a single intensity template. The corresponding label map is computed as in \textit{AtlasMorph-no-Seg}.
\end{itemize}

\section{Results}
Figures \ref{fig:3D_cond_img} and \ref{fig:3D_cond_seg} show visualizations for templates produced by \textit{AtlasMorph} and the \textit{\textit{AtlasMorph}-Uncond} variants. Supplemental Figure~\ref{fig:baseline:3D_ants_uncond},  and Figure~\ref{fig:baseline:3D_ants_cond} show templates optimized using ANTs. The Aladdin templates are shown in Supplementary Figure~\ref{fig:viz:aladinsingle} and Figure~\ref{fig:viz:aladinfrag}. The learned \textit{AtlasMorph} templates are significantly sharper than the average of the population, shown in Figure \ref{fig:exp:3Daverage}. 

Conditional \textit{AtlasMorph} captures variability across age that a single unconditional template cannot. For example, younger subjects have more grey matter, and ventricles get larger as subjects grow older. The intensities of the atlas involve contrast changes that are consistent with segmentation boundaries, capturing the alignment between the intensity templates and the segmentation templates. We include additional visualizations of templates conditioned on age, sex and disease (cognitively normal - mild cognitive impairment - Alzheimer's disease) in the Supplemental Material, Figure \ref{fig:AD_templates}.

\begin{figure*}
\begin{minipage}[t]{\textwidth}

    \includegraphics[width=0.47\textwidth]{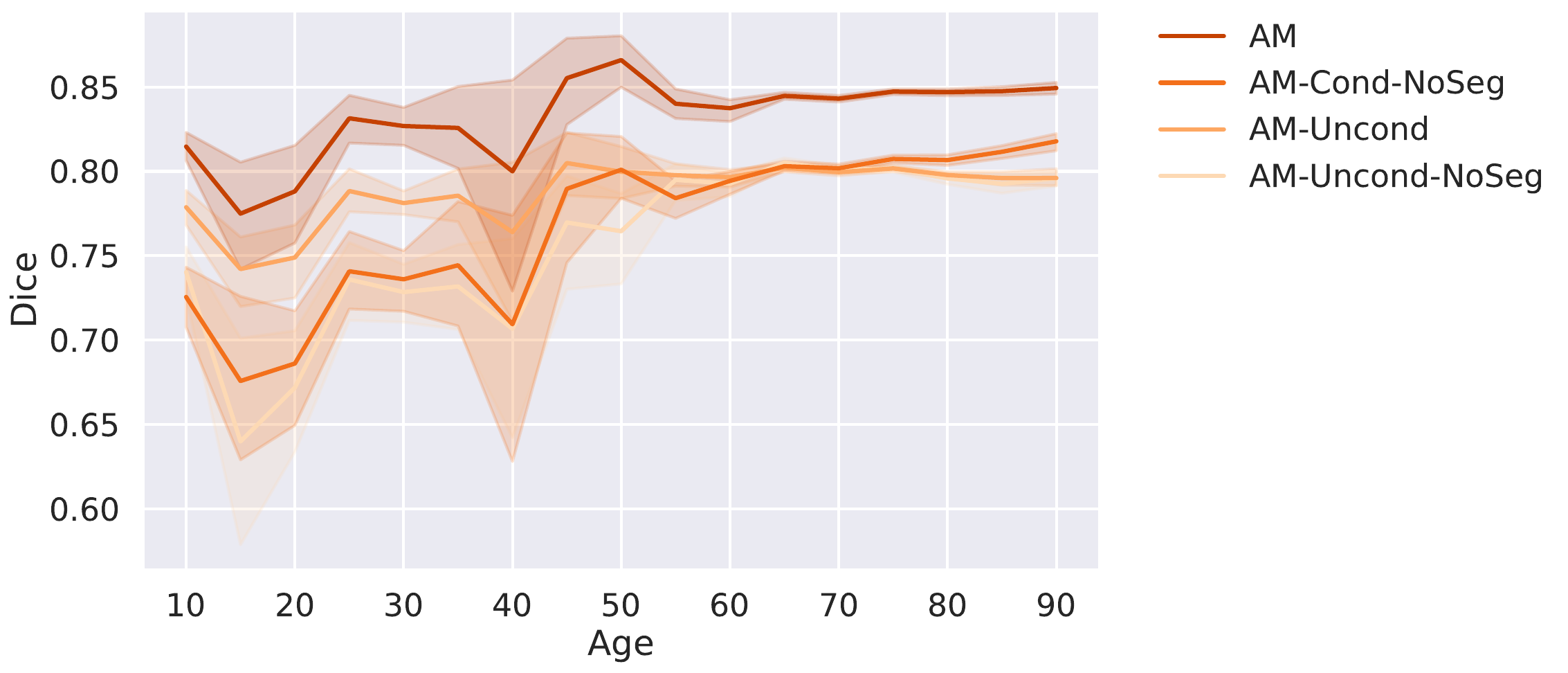}\hfill
    \centering
    \includegraphics[width=0.52\textwidth]{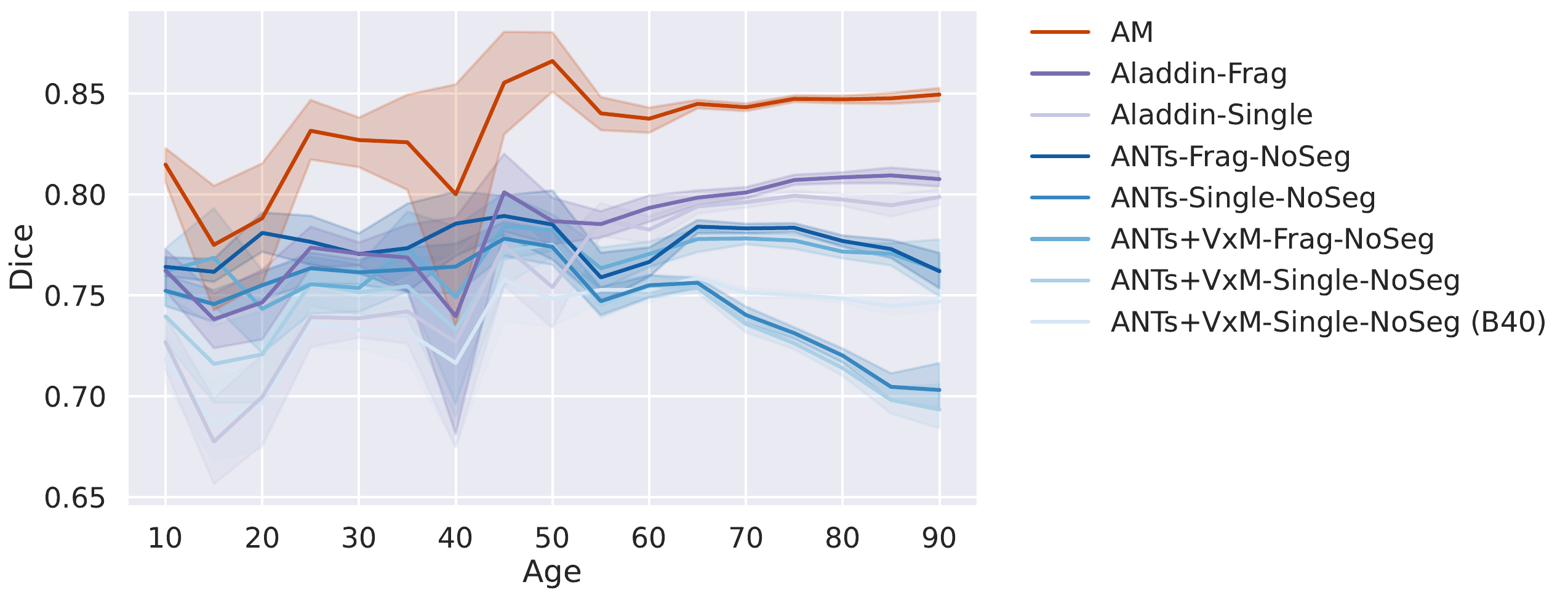}
    \caption{\textbf{Dice score evaluated by age for the test data, higher is better.} The shaded areas are the standard deviation across scans for each age. (Top): \textit{AtlasMorph} compared to ablations variants. (Bottom): \textit{AtlasMorph} compared to the other baselines.}
    \label{fig:3D_age}
\end{minipage}
\end{figure*}

\begin{figure}[t]
  \begin{minipage}[t]{0.48\textwidth}
    \centering
    \includegraphics[width=0.49\textwidth]{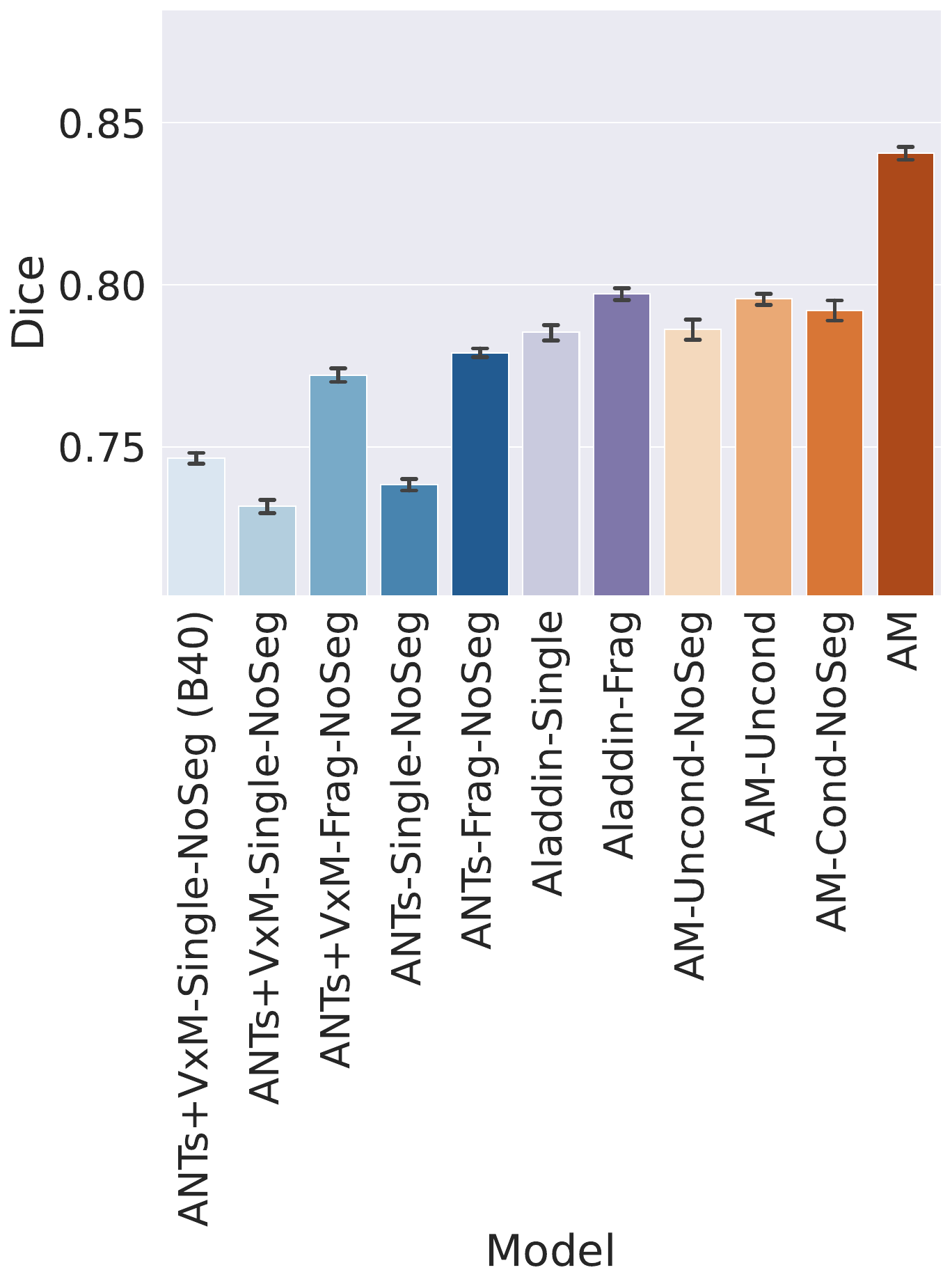}
    \hfill
    \includegraphics[width=0.47\textwidth]{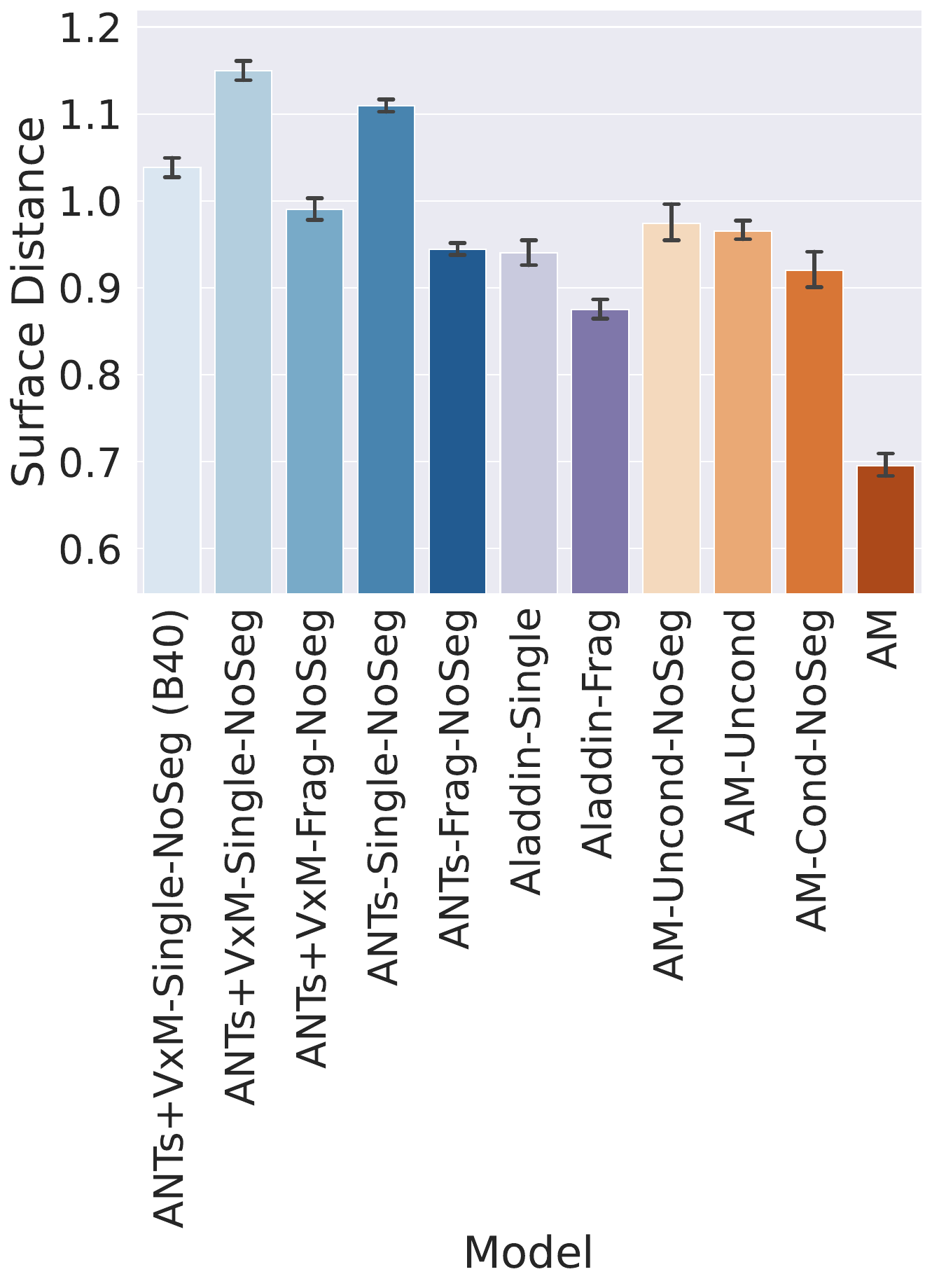}
    \caption{\textbf{Average Dice Score (left, higher is better) and Surface Distance (right, lower is better) for the test set.} The error bars correspond to a 95\% confidence interval.}
    \label{fig:3D_av_Dice_ssd}
  \end{minipage}
  \begin{minipage}[t]{0.49\textwidth}
    \centering
    \hfill
\includegraphics[width=.44\textwidth]{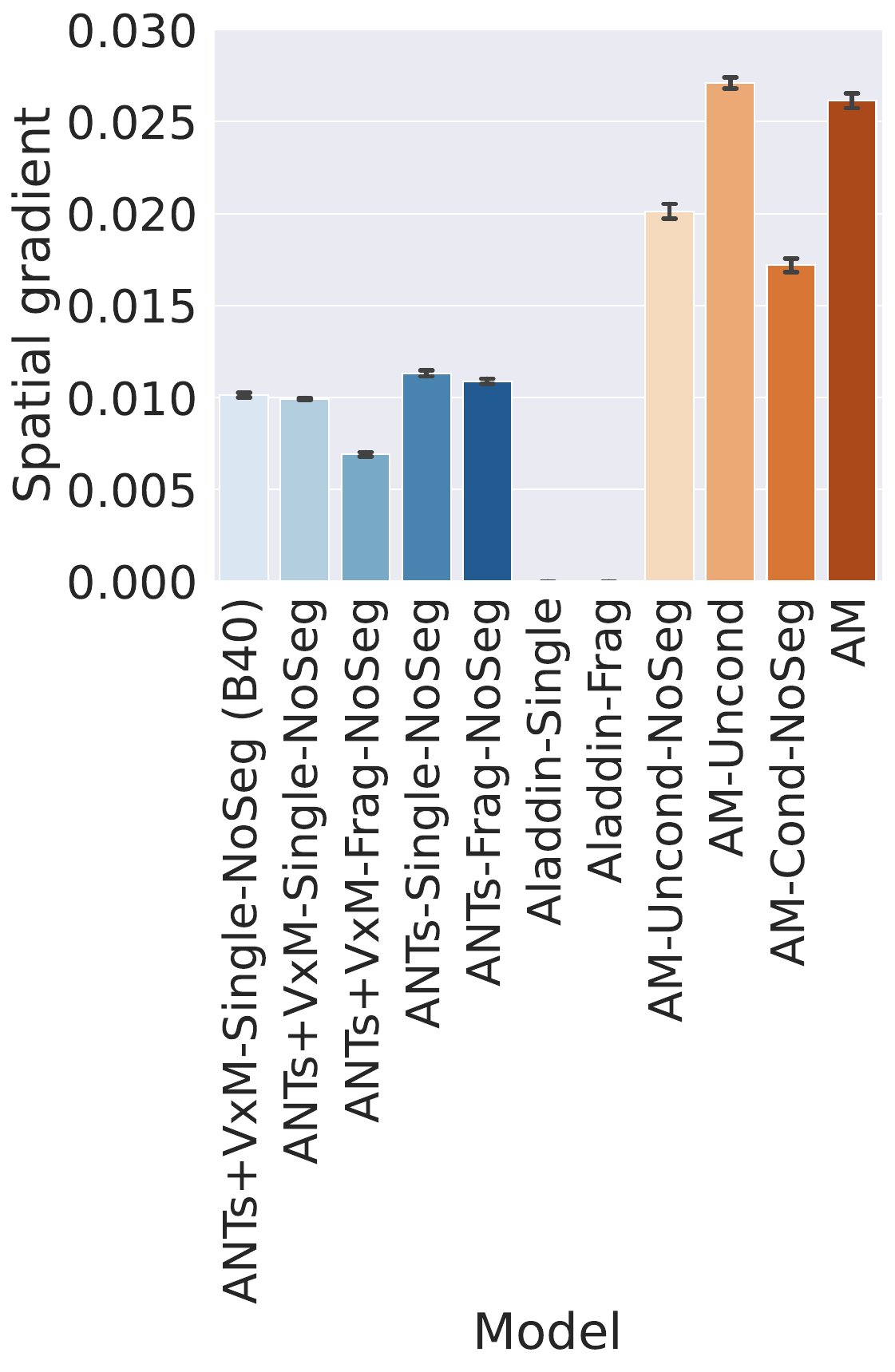}
\includegraphics[width=.49\textwidth]{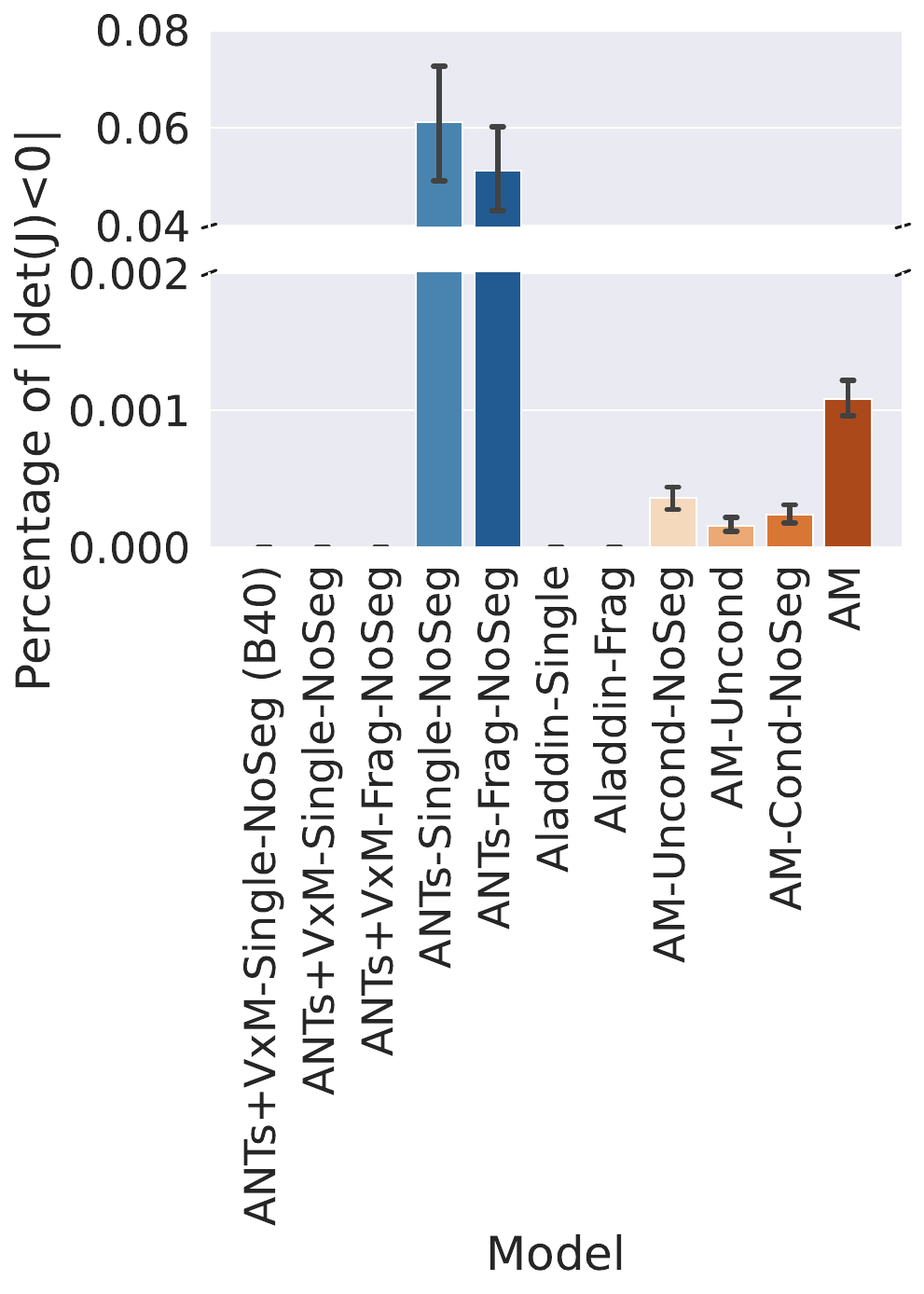}
    \caption{\textbf{Evaluation of smoothness (left) and regularity (right) of deformation fields.} We evaluate the average norm of the gradient of the deformation field, and the percentage of negative values in the Jacobian determinant of the deformation field. The error bar corresponds to the 95\% confidence interval. Lower is better.}
    \label{fig:3D_smoothness_regularity}
  \end{minipage}
\end{figure}

\subsection{Performance on registration}
 
 \noindent\textbf{Evaluation.} We use registration to evaluate the ability of the template to map spatial locations to meaningful anatomical regions. With the obtained deformation fields, we propagate template segmentations to each subject and compare to the manual labels. We evaluate registration segmentation by first measuring the alignment of underlying maps using Dice score and surface distance. To assess the regularity of the deformation field, we count the number of negative values in the Jacobian determinant of the deformation field and evaluate its smoothness, using its spatial gradient. For \textit{ANTs-Single-no-Seg}, we present results for the best of the three seeds we used to compute the template but show individual results in the supplemental material.

\noindent\textit{ a) Average performance.}
For each model variant, we present the average Dice score and surface distance on the test set, in Figure \ref{fig:3D_av_Dice_ssd}. \textit{AtlasMorph} models consistently outperform the registration baselines. Learning label maps jointly with the intensity template significantly improve performances. The conditional variant performs slightly better than the unconditional one for both cases. Fragmented ANTs perform similarly to the unconditional ablation of \textit{AtlasMorph}. We suspect that this gap in performance between the fragmented ANTs methods and our conditional templates is in part due to a lack of data when building the ANTs templates. 

\noindent\textit{b) Per Age Performance.}
We also evaluate the Dice score of \textit{AtlasMorph} model variants when conditioned on ages in Figure \ref{fig:3D_age}. \textit{AtlasMorph} performs best across all ages. Across all models, segmentation of younger subjects is poorer, with the largest standard deviation, and the gap between models is larger. We suspect that this is caused by the large imbalance of subjects in our data: there are more examples of older subjects than younger ones. 

\noindent\textit{ c) Deformation Regularity.} 
Figure \ref{fig:3D_smoothness_regularity} displays the regularity of the deformation fields for all models. For all models, the deformation fields are regular. The very few negative values in the Jacobian determinant, which indicate the fields are not quite diffeomorphic, are caused by numerical errors in the Scaling and Squaring integration. The issue can be alleviated by increasing the integration steps.

We find a higher number of pixel inversions for our \textit{AtlasMorph} model than for other models: 0.001\%. We use the smoothness and regularity metrics as an indication that the deformation fields produced by the network are realistic and interpretable. We used these metrics for hyperparameter tuning when training our networks. 

\begin{figure}[t]
    \centering
    \includegraphics[width=\linewidth]{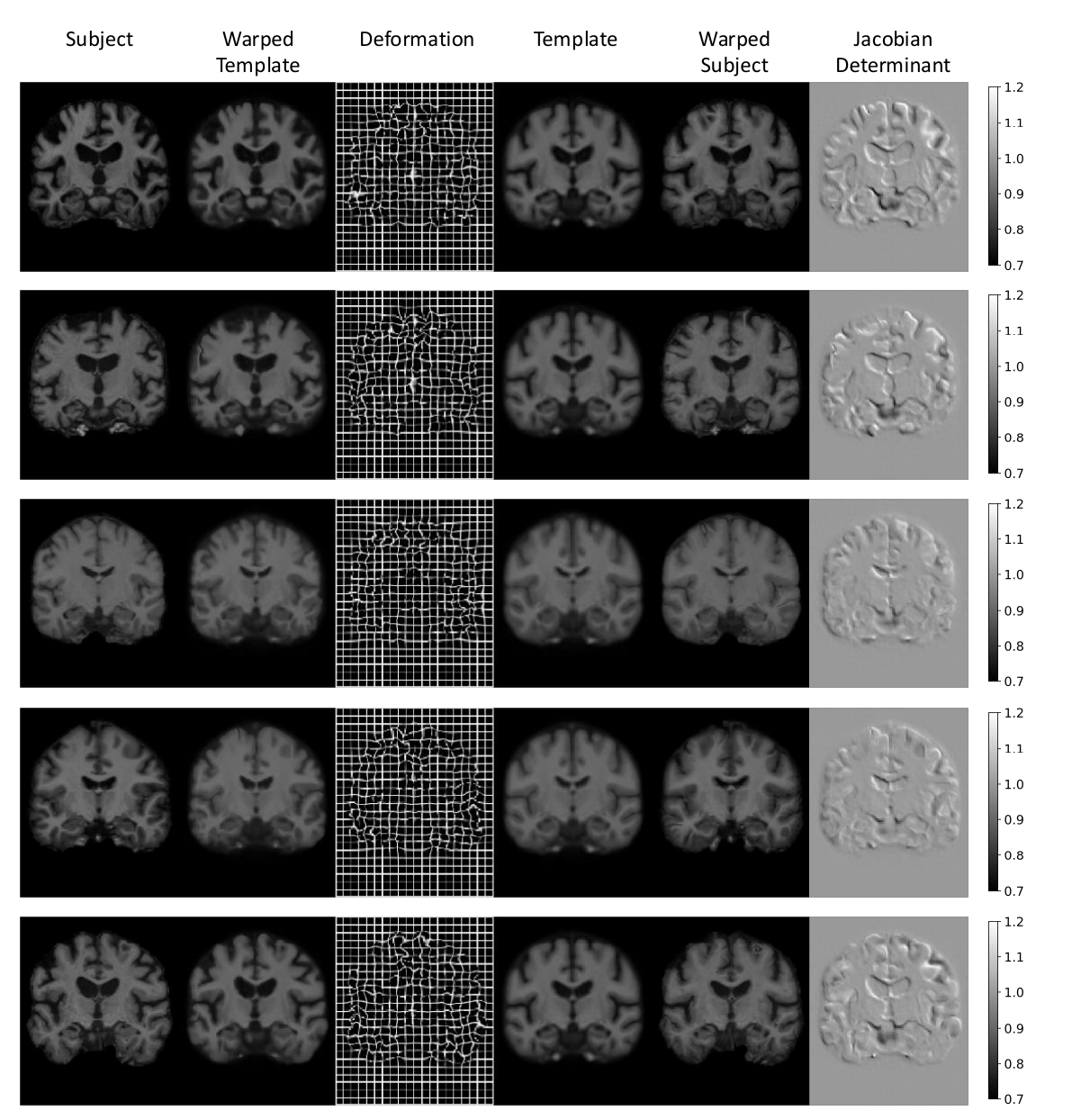}
    \caption{\textbf{Registration example for 5 random subjects.} From left to right: the subject,  the template warped to the subject, the deformation field, the template, the subject warped to the template and the Jacobian determinant. The template is conditional on the age and sex of the subject.}
    \label{fig:exp:registrationexample}
\end{figure}

Figure~\ref{fig:exp:registrationexample} shows example predictions produced by \textit{AtlasMorph} and their corresponding deformation fields and Jacobian determinants. The values of the Jacobian determinant are close to 1, which indicates small deformation norms. 

\subsection{Analysis}
We aim to find central, unbiased templates for given attributes. We employ the centrality term in our loss function. To avoid only using that same measure to evaluate the “centrality” of the result, we also assess if the produced conditional template captures population trends.
We first analyze the ability of the proposed template functions to capture population trends. We then use this analysis to assess the value of the new centrality formulation, by comparing our template function to a template function learned using the global centrality definition $\bar{\bu} = \sum_{k=1}^K \bu_{k}$, introduced in~\cite{dalca2019learning}. To achieve this, we train a model with the same setup and hyperparameters as \textit{AtlasMorph}, but replace the proposed centrality term with the one used in~\cite{dalca2019learning}. The only difference between \textit{AtlasMorph} and this model variant is the way the centrality is computed. We call this model Learnable Templates 2019, abbreviated LT2019.

We approximate the \textit{ground truth population trend} by computing the volumes from manual segmentations of all scans, and applying kernel density estimation (KDE) with bandwidth of 5 to estimate the population average at each age. We then evaluate the quality of label map volumes captured by each method, \textit{AtlasMorph} and LT2019, by computing, for each age and sex, the relative error between the manual and predicted volumes. For central templates, the corresponding structure volumes should match the population structures computed using KDE.

We show results on the test set and include training data results in the supplemental material.

\begin{figure*}
  \begin{minipage}[t]{\textwidth}
    \centering
        \includegraphics[width=0.49\textwidth]{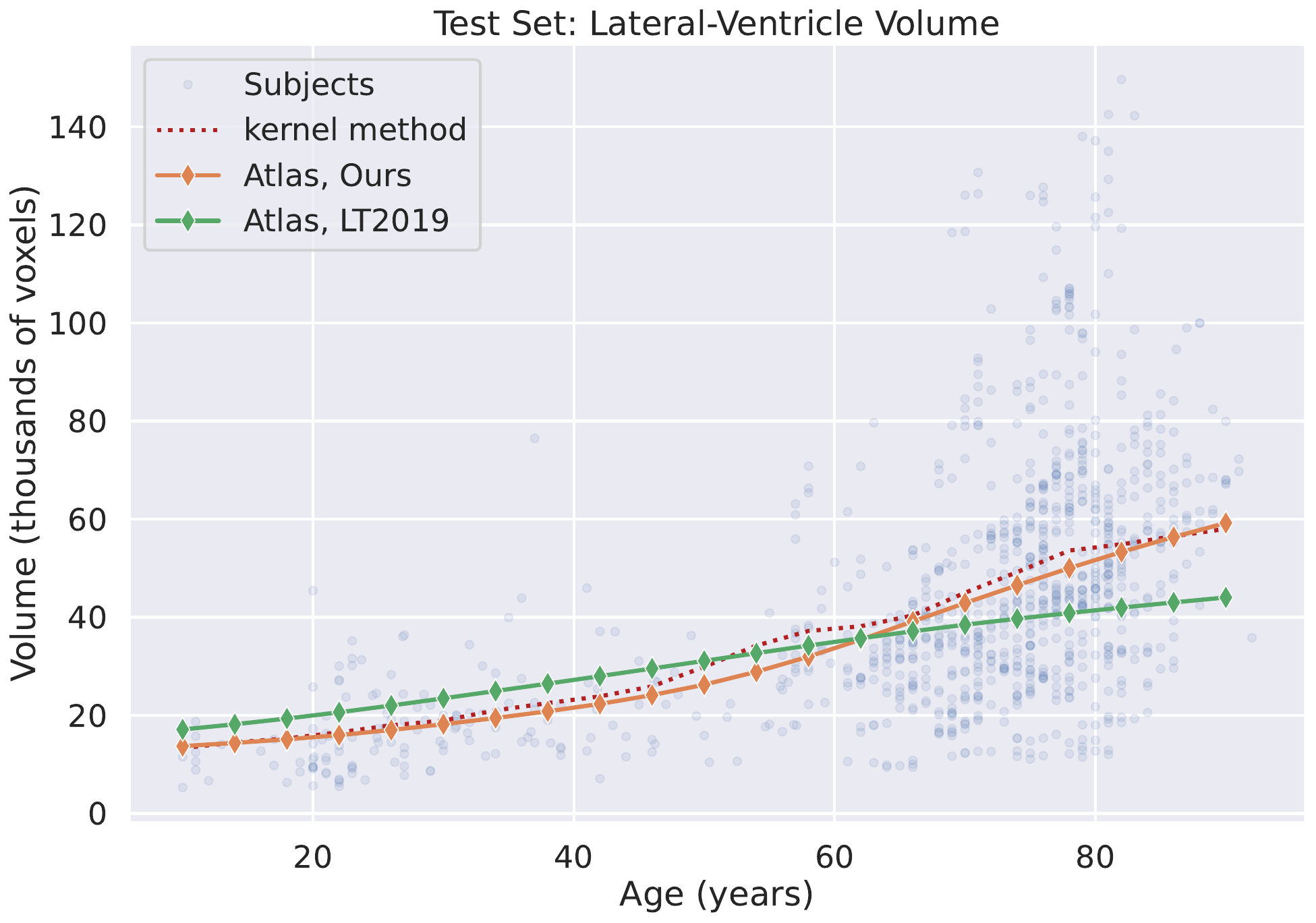} \hfill
    \includegraphics[width=0.49\textwidth]{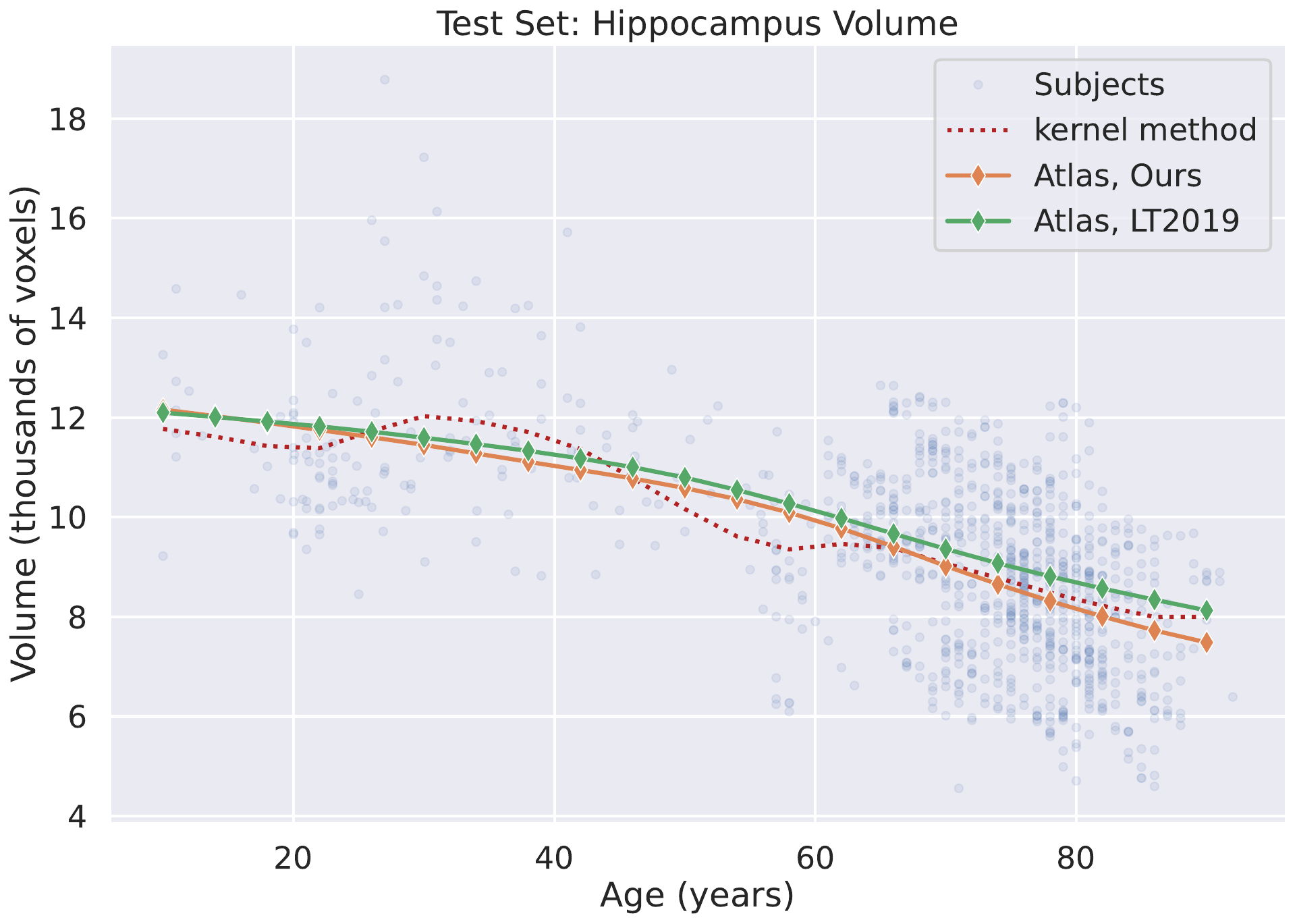}
    \caption{\textbf{Centrality using ventricles and hippocampi volumes conditioned on age for the test set.} The blue dots show the structure size (ventricles on the left, hippocampi on the right) for each brain MRI, sorted by the age of the subject when the MRI was taken. The starred orange line represents the volume captured by \textit{AtlasMorph}. The volume captured by a kernel method is shown in red and the one captured by LT2019~\cite{dalca2019learning} in green. \textit{AtlasMorph} captures more closely the population than LT2019.}
    \label{fig:3D_vent_vol_kernel}
  \end{minipage}
  \end{figure*}

\begin{figure}[!ht]
\centering
\includegraphics[width=0.47\textwidth]{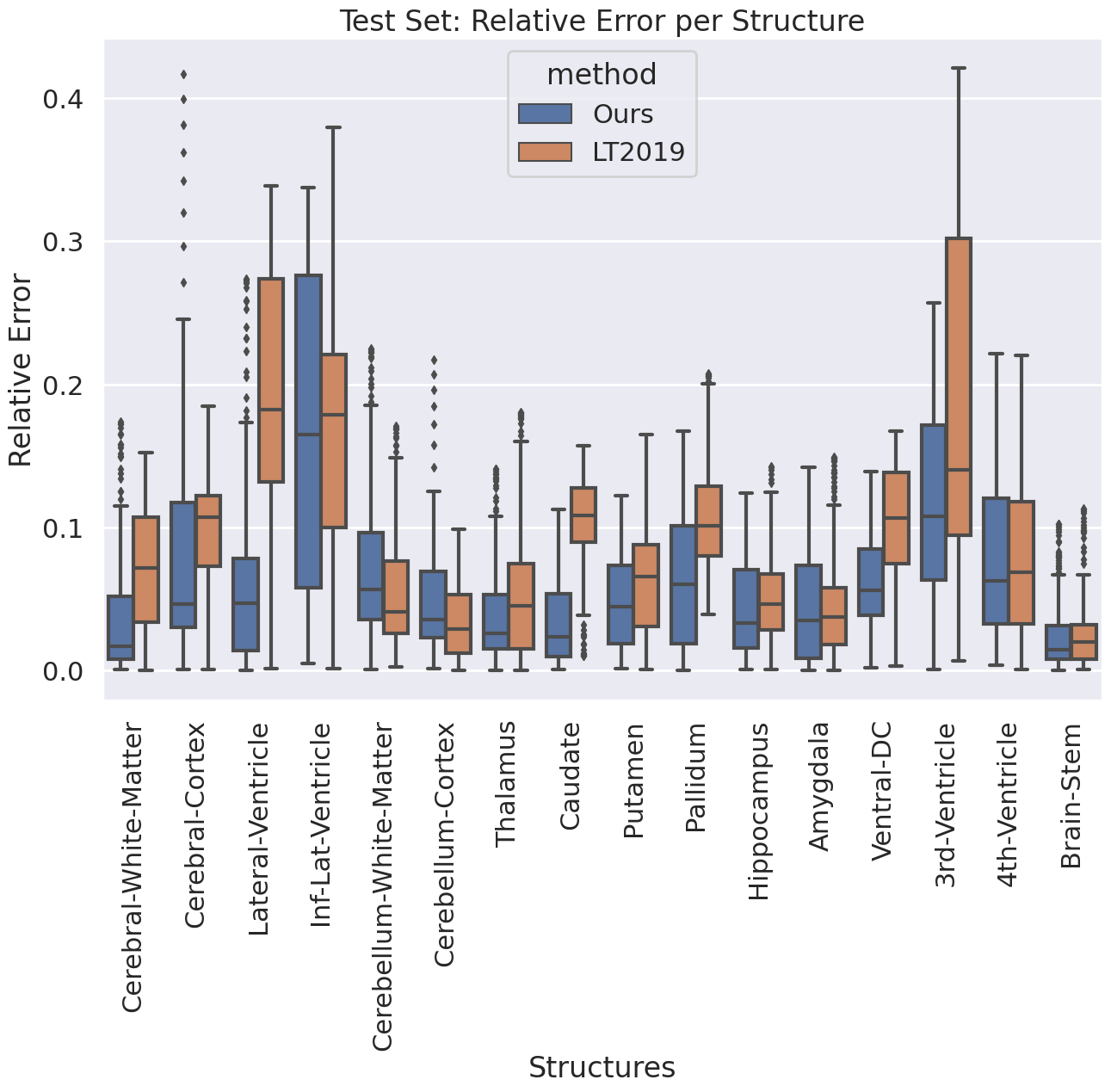}
\caption{\textbf{Comparison of the proposed centrality model with the baseline (unweighted) version for all structures.} We compute the relative error between structure volumes from \textit{AtlasMorph} templates and a kernel based estimation for both centrality definitions, and show the distribution of relative error per structure for the test set. \textit{AtlasMorph} leads to templates closer to the population.}
\label{fig:3D_strct_vol}
\end{figure}

\vspace{0.1cm}\noindent\textbf{Results.}
We first focus on ventricles and hippocampi, two structures that typically change noticeably with age. As shown visually in Figures \ref{fig:3D_cond_img} and \ref{fig:3D_cond_seg}, and quantitatively in Figure \ref{fig:3D_vent_vol_kernel}, the increase of ventricles and a shrinkage of hippocampi in the templates reproduce known trends in a normally aging population. Figure \ref{fig:3D_vent_vol_kernel} shows evidence that \textit{AtlasMorph}, driven by the new centrality loss, captures the population trend, matching the kernel density estimation better than LT2019, for both ventricles and hippocampi structures. Figure \ref{fig:3D_strct_vol} shows that this result holds for most other structures as well. \textit{AtlasMorph} gives structure volumes close to the kernel based method estimates. We present population trends visualizations for additional structures in the supplemental material.

%
%

\subsection{Influence of initialization}
We run additional analysis to evaluate the influence of different initializations on the learned templates~\cite{avants2004geodesic, joshi2004unbiased}. We train \textit{AtlasMorph} in two settings.
\begin{itemize}
    \item Initializing with a single randomly chosen subject: We choose three random subjects and train one network per random subject. 
    \item Taking the mean of 100 random subjects: We repeat this experiment twice with different sets of 100 subjects. 
\end{itemize}

\begin{figure}
    \centering
    \includegraphics[width=0.49\linewidth]{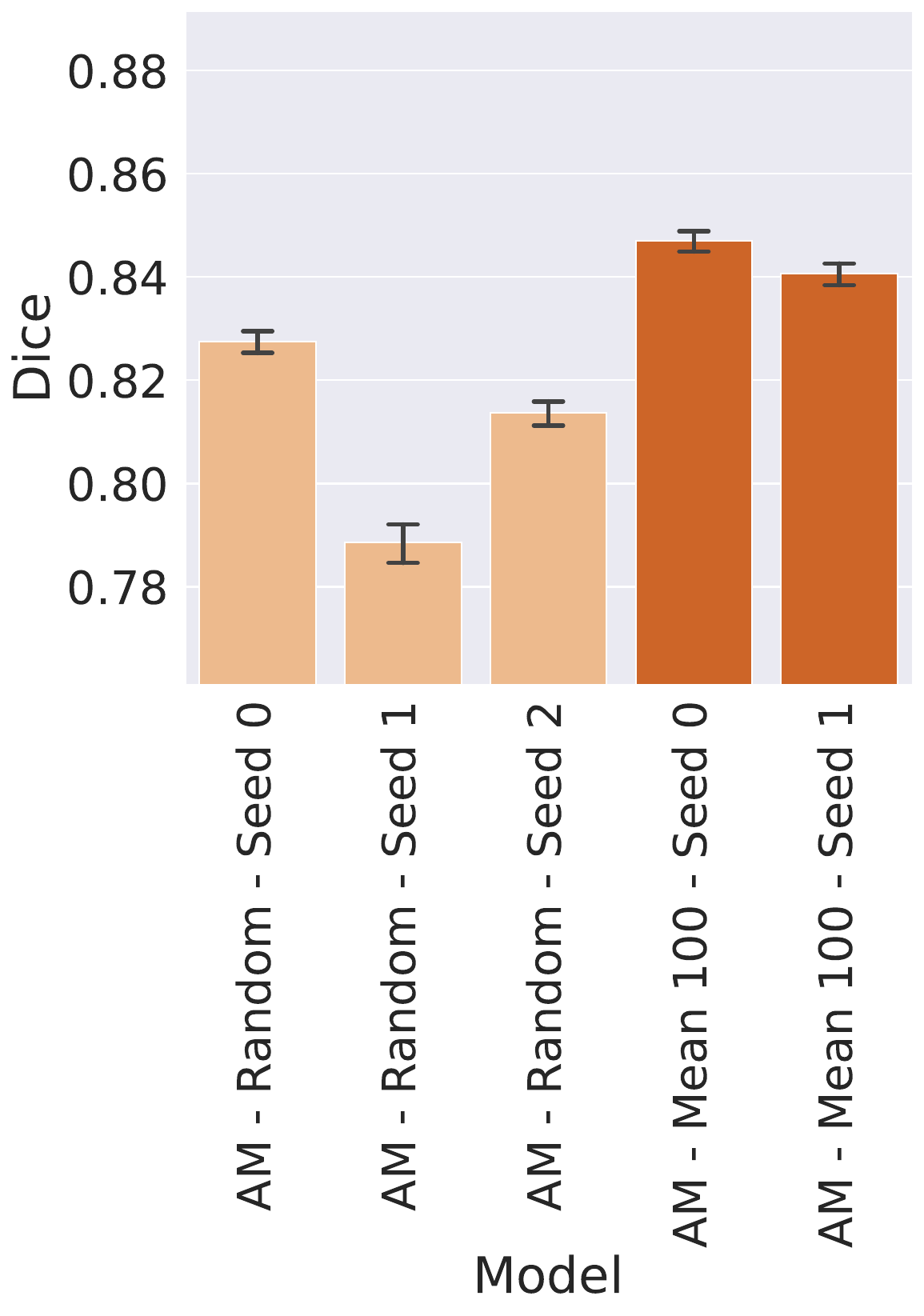}
    \hfill
    \includegraphics[width=0.48\linewidth]{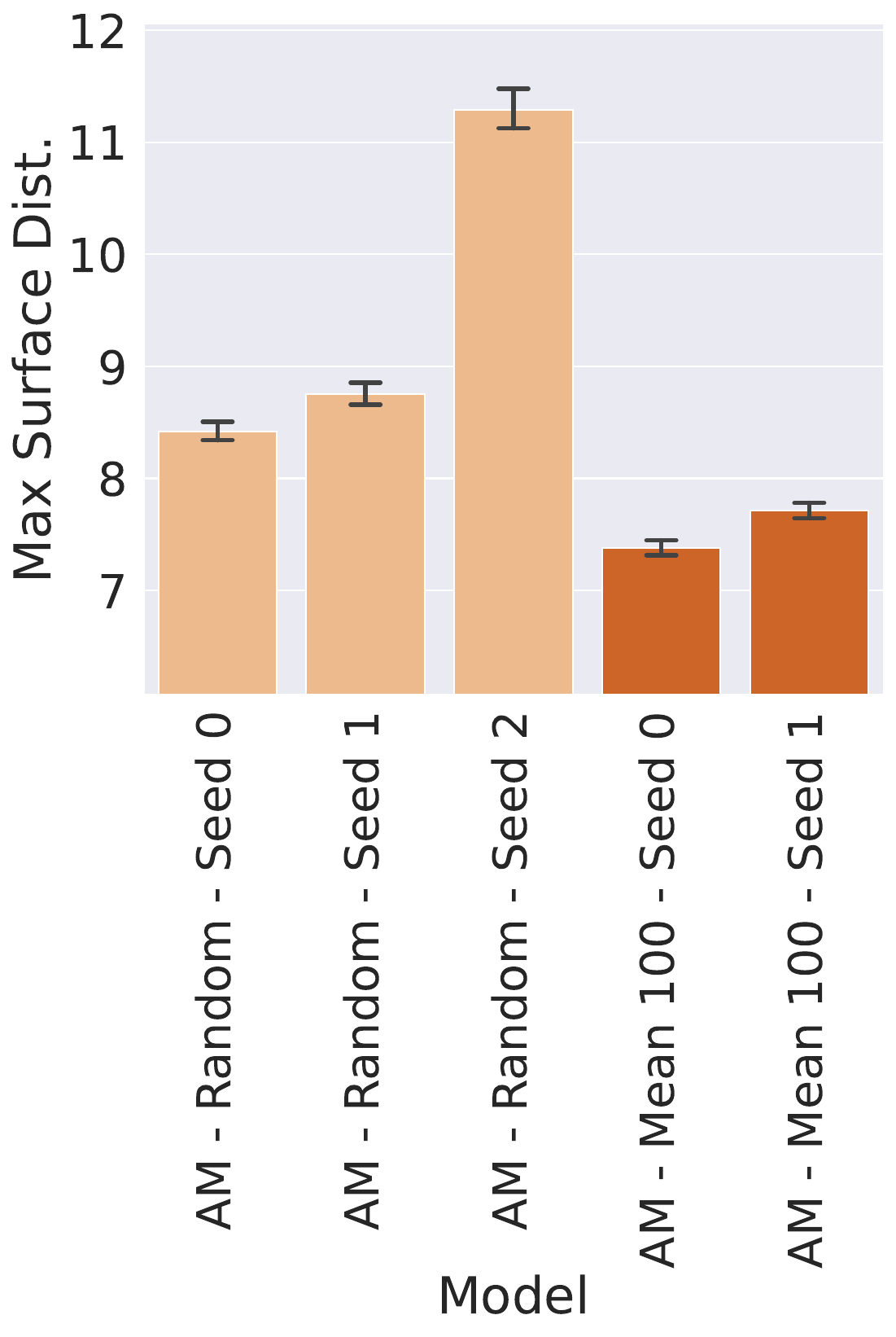}
    \caption{\textbf{Influence of different \textit{AtlasMorph} initializations.} With respect to both Dice score (left) and surface distance (right), initializing with the mean of multiple subjects yields better and more stable results than initializing with a single subject.}
    \label{fig:exp:bias}
\end{figure}

\vspace{0.1cm}\noindent\textbf{Results.} Figure \ref{fig:exp:bias} shows that selecting a random subject leads to highly variable and worse performance. Models initialized with the mean of 100 subjects yield better results with lower variance compared to the models initialized with a single subject. We recommend initializing our template with a population average for better performance.

\section{Limitations} 
In this work, we focused on a limited number of attributes: mostly age and sex. These attributes are widely available across datasets. We also used the disease diagnosis attribute, analyzed in the supplementary material, available only for ADNI and OASIS. However, various other attributes can be included, especially those known or suspected to play a role in brain volumetric changes or brain-related diseases, such as APOE genetic variants. These additional attributes present new challenges, such as balancing their impact on the produced template, especially when jointly using a large number of attributes. 

Because of prohibitive run times, it was not possible for us to build multiple ANTs templates involving all data available, and we used subsets of 100 representative subjects for each separate template. While we believe that the ANTs performance in this paper is representative, it is possible that including more subjects would lead to minor improvements.

\section{Conclusion}
We showed that \textit{AtlasMorph} can learn realistic and representative conditional templates with corresponding label maps on brain anatomy, and captures important population trends. \textit{AtlasMorph} captures important population trends such as brain atrophy with age that can aid population studies and disease diagnosis. We find that the new conditional centrality formulation leads to templates more representative of subpopulations with \textit{AtlasMorph} than with previous work. 

We learn conditional label maps and show that this substantially improves segmentation performance, while eliminating the need for post-processing alignment. \textit{AtlasMorph} outperforms the baselines, as well as the \textit{AtlasMorph} ablations. 

In future research, we plan to investigate the potential of this framework under distribution shift, including different acquisition scenarios, different data collection protocols, low quality clinical data, or limited data. In general, templates are computed for a specific anatomy (brain in this work), often for specific modality (T1-weighted MRI in this work).  
This can greatly limit analysis, and practitioners cannot use potentially useful data. We plan to expand on recent synthesis-based methods~\cite{billot2021synthseg,hoffmann2021synthmorph,hoopes2022synthstrip} and non parametric mechanisms~\cite{butoi2023universeg,rakic2024tyche} to build a modality invariant template: given scans from an anatomy, the template could be used irrespective of the scan's modality.

\textit{AtlasMorph} enables one to efficiently build conditional templates quickly. This is in contrast to classical methods, which are often too slow to obtain templates for every dataset or analysis. We hope that \textit{AtlasMorph} will lead to a paradigm shift in which rapid generation of templates and the corresponding label maps will greatly enhance registration quality and population characterization for medical imaging research studies.

\section*{Acknowledgments}
Research reported in this paper was supported by the National Institute of Biomedical Imaging and Bioengineering of the National Institutes of Health under award number R01EB033773, R01AG053949 and 5R01AG064027. This material is based upon work supported by the National Science Foundation under Grant No. NSF CAREER 1748377. Finally, this work was also supported in part by funding from the Eric and Wendy Schmidt Center at the Broad Institute of MIT and Harvard as well as Quanta Computer Inc. Finally, some of the computation resources required for this research was performed on computational hardware generously provided by the Massachusetts Life Sciences Center.

\section*{Acknowledgments}
Research reported in this paper was supported by the National Institute of Biomedical Imaging and Bioengineering of the National Institutes of Health under award number R01EB033773, R01AG053949 and 5R01AG064027. This material is based upon work supported by the National Science Foundation under Grant No. NSF CAREER 1748377. Finally, this work was also supported in part by funding from the Eric and Wendy Schmidt Center at the Broad Institute of MIT and Harvard as well as Quanta Computer Inc. Finally, some of the computation resources required for this research was performed on computational hardware generously provided by the Massachusetts Life Sciences Center.

Data collection and sharing for this project was funded by the Alzheimer's Disease Neuroimaging Initiative (ADNI) (National Institutes of Health Grant U01 AG024904) and DOD ADNI (Department of Defense award number W81XWH-12-2-0012). ADNI is funded by the National Institute on Aging, the National Institute of Biomedical Imaging and Bioengineering, and through generous contributions from the following: AbbVie, Alzheimer’s Association; Alzheimer’s Drug Discovery Foundation; Araclon Biotech; BioClinica, Inc.; Biogen; Bristol-Myers Squibb Company; CereSpir, Inc.; Cogstate; Eisai Inc.; Elan Pharmaceuticals, Inc.; Eli Lilly and Company; EuroImmun; F. Hoffmann-La Roche Ltd and its affiliated company Genentech, Inc.; Fujirebio; GE Healthcare; IXICO Ltd.; Janssen Alzheimer Immunotherapy Research \& Development, LLC.; Johnson \& Johnson Pharmaceutical Research \& Development LLC.; Lumosity; Lundbeck; Merck \& Co., Inc.; Meso Scale Diagnostics, LLC.; NeuroRx Research; Neurotrack Technologies; Novartis Pharmaceuticals Corporation; Pfizer Inc.; Piramal Imaging; Servier; Takeda Pharmaceutical Company; and Transition Therapeutics. The Canadian Institutes of Health Research is providing funds to support ADNI clinical sites in Canada. Private sector contributions are facilitated by the Foundation for the National Institutes of Health (www.fnih.org). The grantee organization is the Northern California Institute for Research and Education, and the study is coordinated by the Alzheimer’s Therapeutic Research Institute at the University of Southern California. ADNI data are disseminated by the Laboratory for Neuro-Imaging at the University of Southern California.


{
    \small
    \bibliographystyle{ieeenat_fullname}
    \bibliography{bibliography}
}

\newpage

\begin{figure}[h!]
    \centering
    \includegraphics[width=0.45\textwidth]{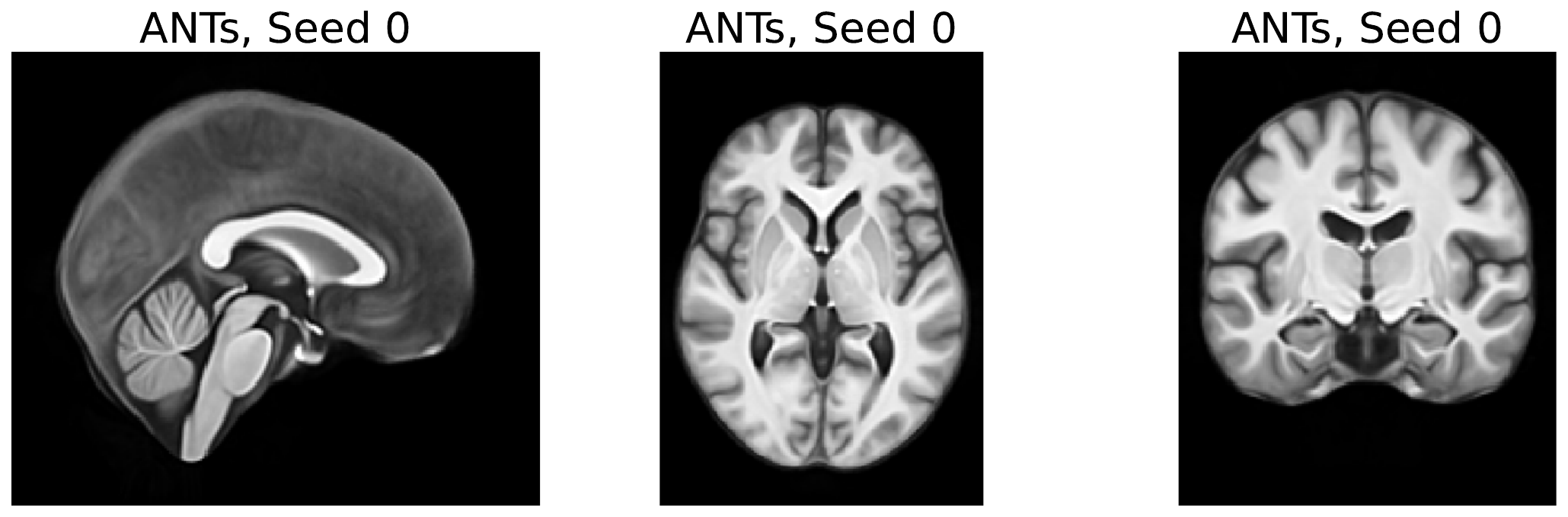}
    \caption{\textbf{\textit{ANTs-Single-no-Seg} iterative template used in our registration baseline.} Shown is a template built using ANTs SyGN, sampling 100 subjects uniformly by age and sex.}
    \label{fig:baseline:3D_ants_uncond}
\end{figure}
\begin{figure}[h!]
    \centering
    \includegraphics[width=0.45\textwidth]{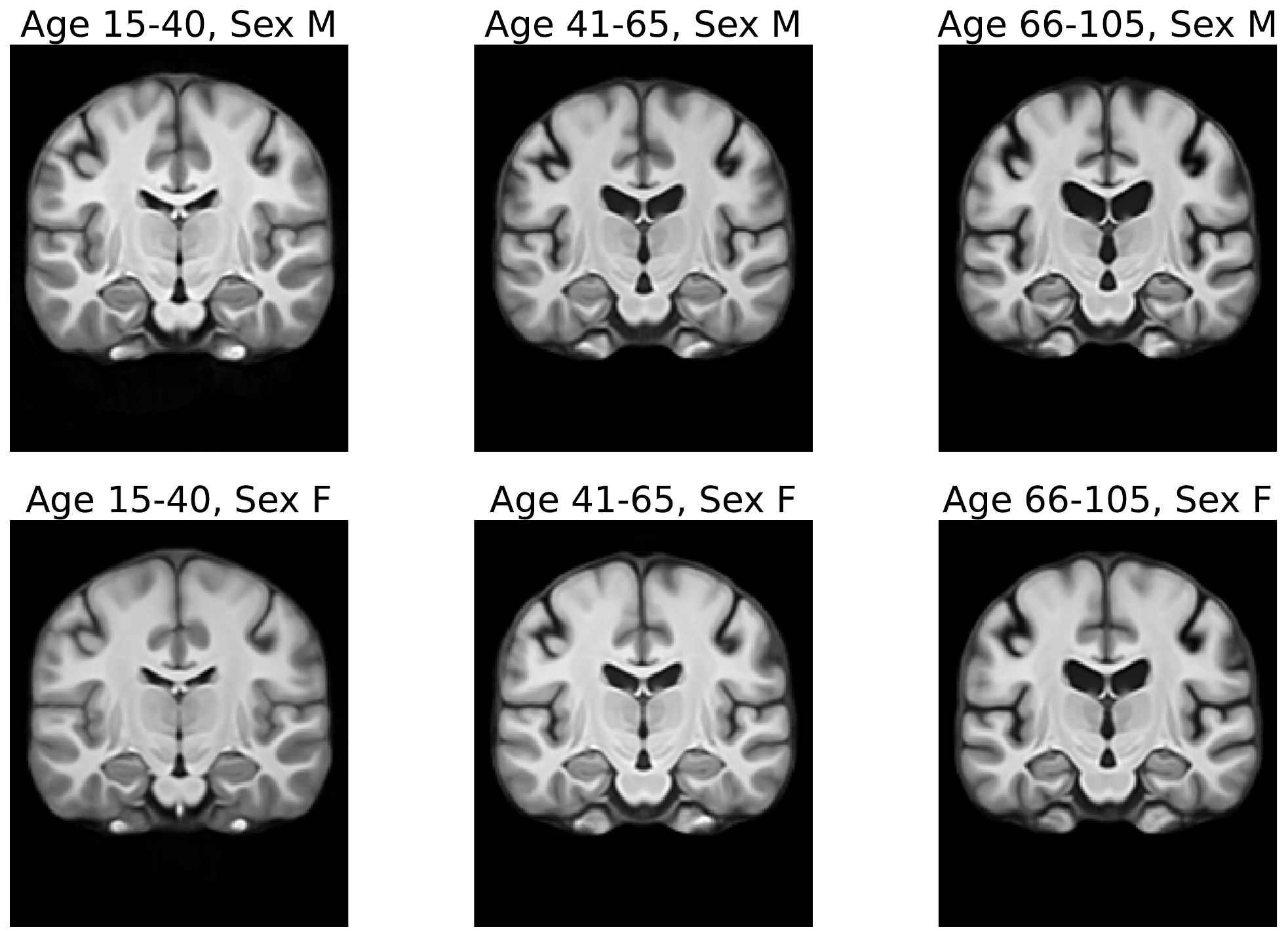}
    \caption{\textbf{\textit{ANTs-Frag-no-Seg} iterative templates used in our registration baseline.} The templates have been build by applying the ANTs SyGN algorithm to six different population subgroup partitioned by age and sex. Each group has 100 subjects.}
    \label{fig:baseline:3D_ants_cond}
\end{figure}

\begin{figure}[h]
    \centering
    \includegraphics[width=\linewidth]{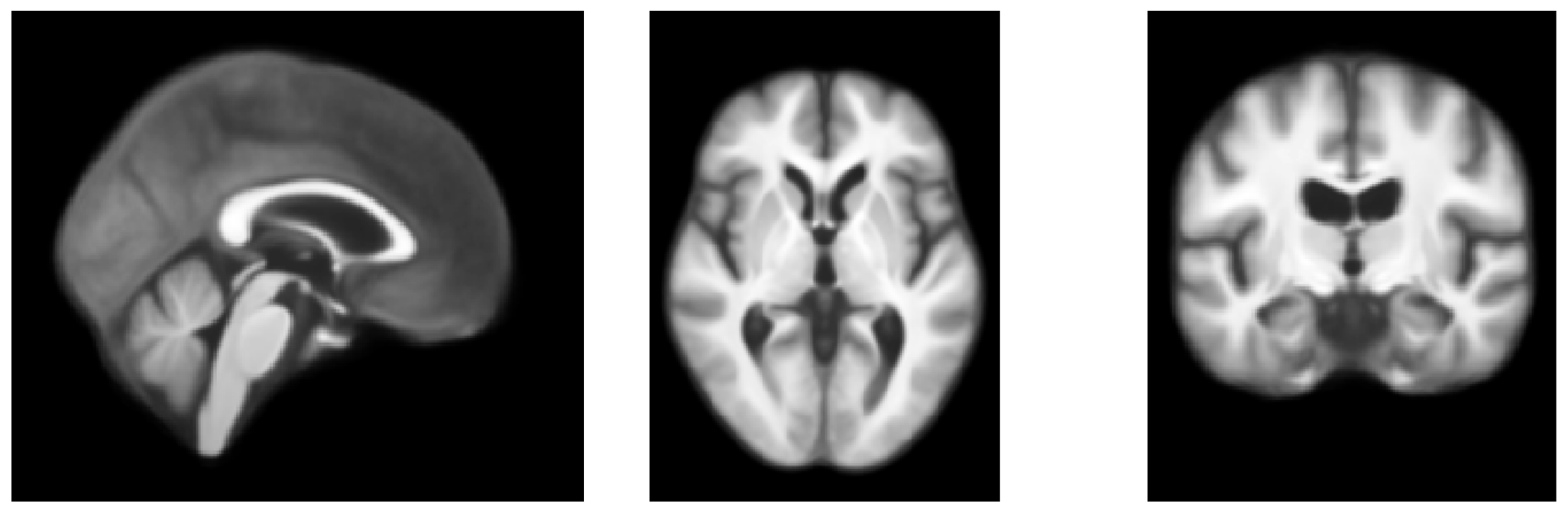}
    \caption{\textbf{\textit{Aladdin-Single} template learned using Aladdin on the entire population.}}
    \label{fig:viz:aladinsingle}
\end{figure}
\begin{figure}[h]
    \centering
    \includegraphics[width=\linewidth]{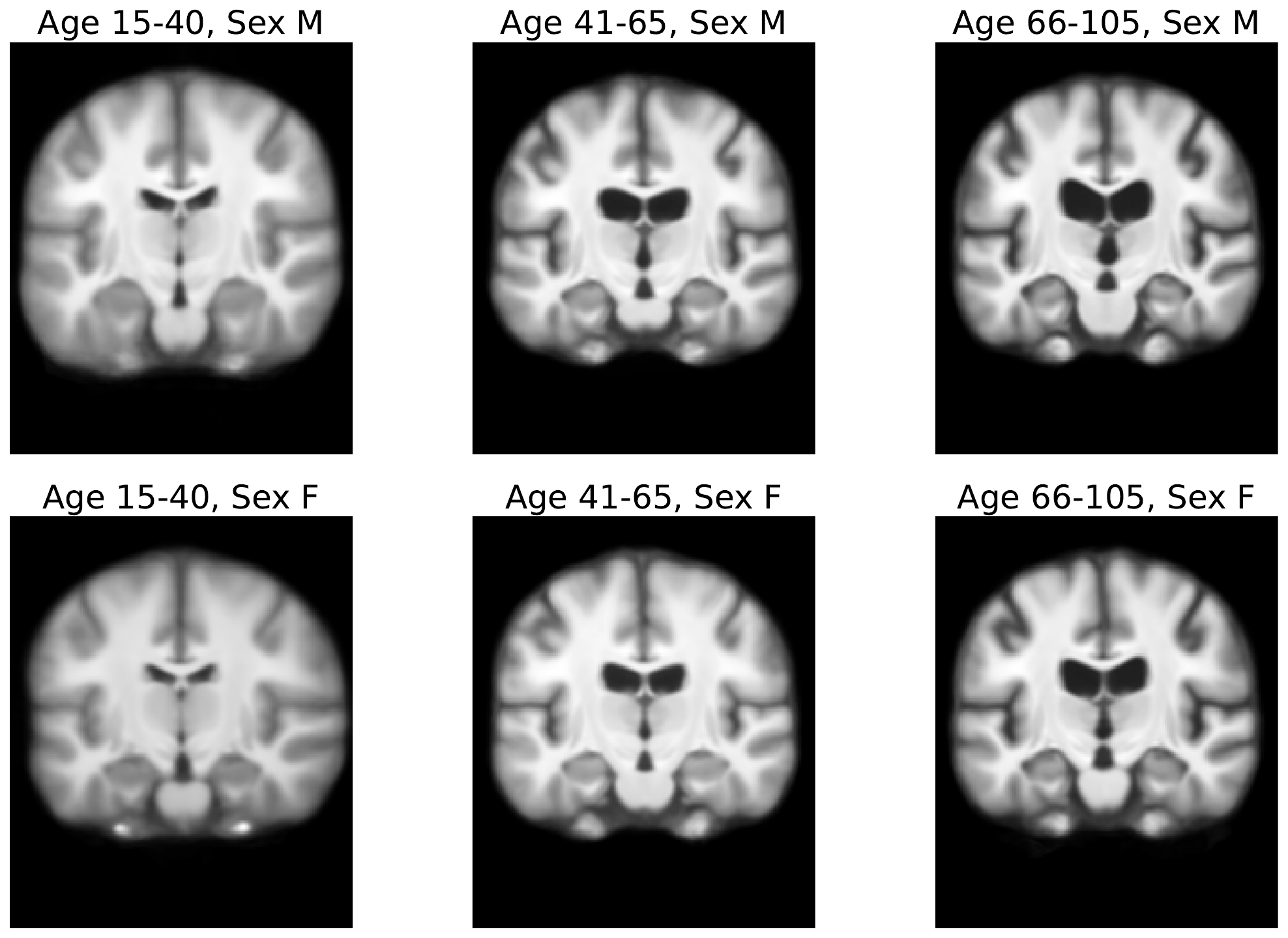}
    \caption{\textbf{\textit{Aladdin-Frag} templates learned using Aladdin after binning the population.} The templates have been learned using one model per population subgroup, partitioned by age and sex.}
    \label{fig:viz:aladinfrag}
\end{figure}

\begin{figure}[h]
    \centering
    \includegraphics[width=0.49\textwidth]{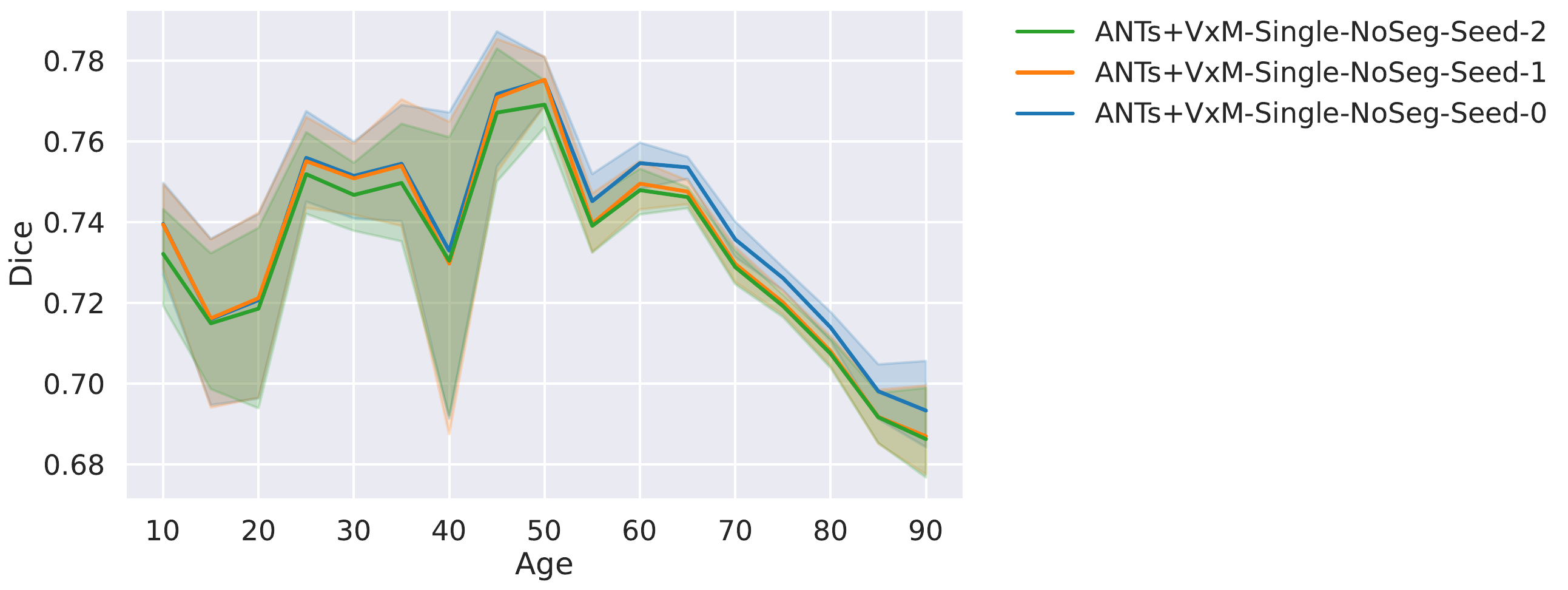}
    \caption{\textbf{Dice score evaluated by age for the different \textit{ANTs+VxM-Single-no-Seg} templates built with different seeds.} For Dice score, higher is better. The shaded areas are the standard deviation for each age.}
    \label{fig:ants_seeds}
\end{figure}
\section*{Supplementary Material}
\subsection{Single ANTs Templates}
\label{s:unconditional-ants}
We build \textit{ANTs-Single-no-Seg} by selecting at random 100 subjects, stratified by age, to ensure an even distribution across ages. We use the default ANTs template construction script with default construction parameters and $4$ template updates. For the pairwise registrations,  we use the ANTs greedy SyGN algorithm with a gradient step of $0.1$, $\sigma_{flow}=0.9$ and $\sigma_{total}=0.2$. We use the squared local normalized cross-correlation with a radius of 3 as our image similarity metric. We equally weigh the image similarity metric and the image transformation metric. We use a four-level registration pyramid, downsampling the images by a factor of [$6$, $4$, $2$, $1$] and smoothing with a Gaussian kernel with standard deviations of [$3$, $2$, $1$, $0$] for each respective stage of the pyramid. The registration is done with a maximum of [$100$, $100$, $70$, $20$] iterations at each respective stage. We do not use bias field correction nor linear registration as they are part of our data-processing pipeline. We build templates for three seeds. The performance by age for each seed is give in Figure \ref{fig:ants_seeds}. All seeds show similar performance, with highest Dice scores for the middle-aged group and lower scores for older and younger subjects. The resulting template is shown in Figure~\ref{fig:baseline:3D_ants_uncond}.
%

%
\subsection{Frangmented ANTs Templates}
\label{s:conditional-ants}
We build the fragmented ANTs templates using the same method as the single ANTs templates. For each of our 6 templates, we sample 100 subjects from each group. Our groups are partitioned into male and female with three age groups each: 15-40, 41-65 and 66-100. The resulting templates are shown in Figure~\ref{fig:baseline:3D_ants_cond}.

\subsection{Aladdin}

We train our Aladdin networks with the following weights for each term of the loss function: 10,000 for the regularization, 10 for the similarity and 0.2 for the image pair. The Aladdin templates are shown in Figure \ref{fig:viz:aladinsingle} and Figure \ref{fig:viz:aladinfrag}. We train an Aladdin network that matches the \textit{AtlasMorph} network in terms of architecture. 
\begin{figure}[t]
\centering
\includegraphics[width=0.47\textwidth]{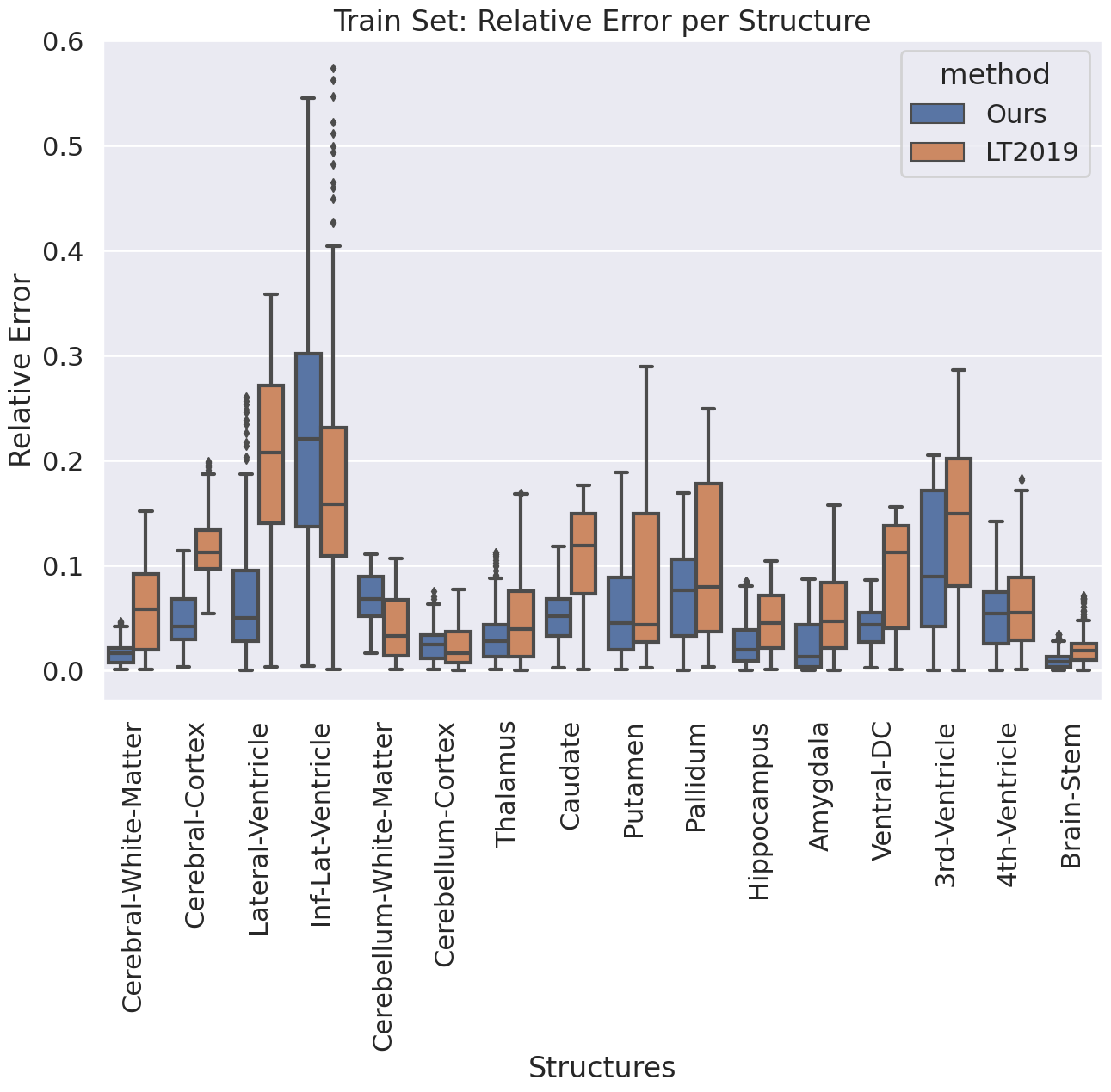}
\caption{\textbf{Comparison of the proposed centrality model with the baseline (unweighted) version for all structures on the training data.} We compute the relative error between ventricle volumes from \textit{AtlasMorph} templates and a kernel based estimation for both centrality definitions, and show the distribution of relative error per structure for the training set. \textit{AtlasMorph} leads to templates closer to the population for each attribute.}
\label{fig:3D_strct_vol_train}
\end{figure}

\subsection{Alzheimer Disease Attribute}
\begin{figure*}[h!]
    \centering
    \includegraphics[width=\textwidth]{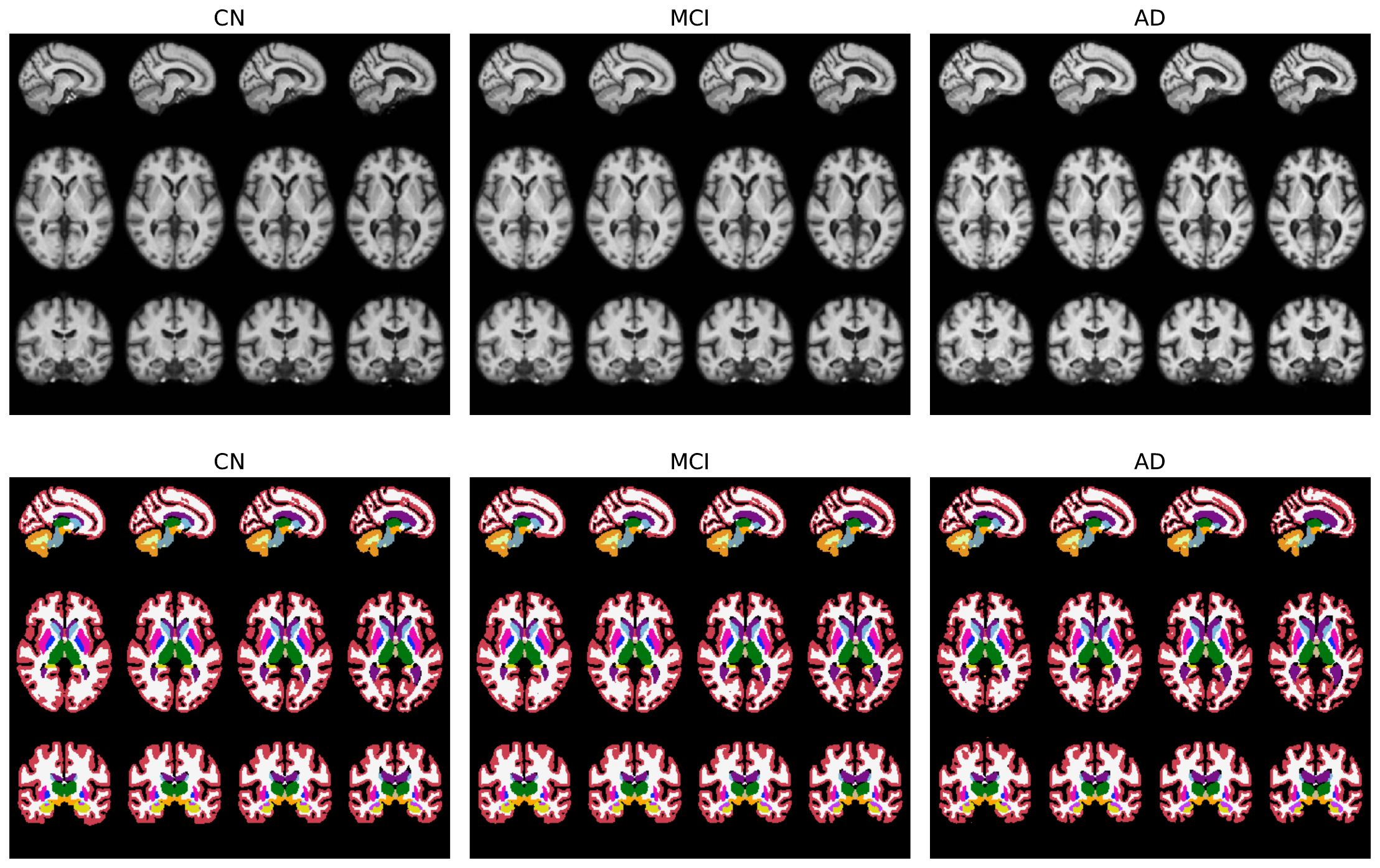}
    \caption{Templates learned based on age, sex and cognitive state attributes with the corresponding segmentation labels. Left to right: cognitively normal, mild cognitive impairment and Alzheimer's disease. For each cognitive group, we show templates for four growing ages (55 to 90) and fix the sex to male.}
    \label{fig:AD_templates}
\end{figure*}

We also learn templates conditioned on age, sex and stages of cognitive impairment: Alzheimer's Disease (AD), Mild Cognitive Impairment (MCI) and Cognitively Normal (CN). We train on the ADNI and OASIS datasets that have a disease label as well as age and sex. The learned template examples are provided in Figure \ref{fig:AD_templates}. We observe that the AD templates present larger ventricles compared to the templates with similar age and sex.

\begin{figure*}
    \centering
    \includegraphics[width=0.23\textwidth]{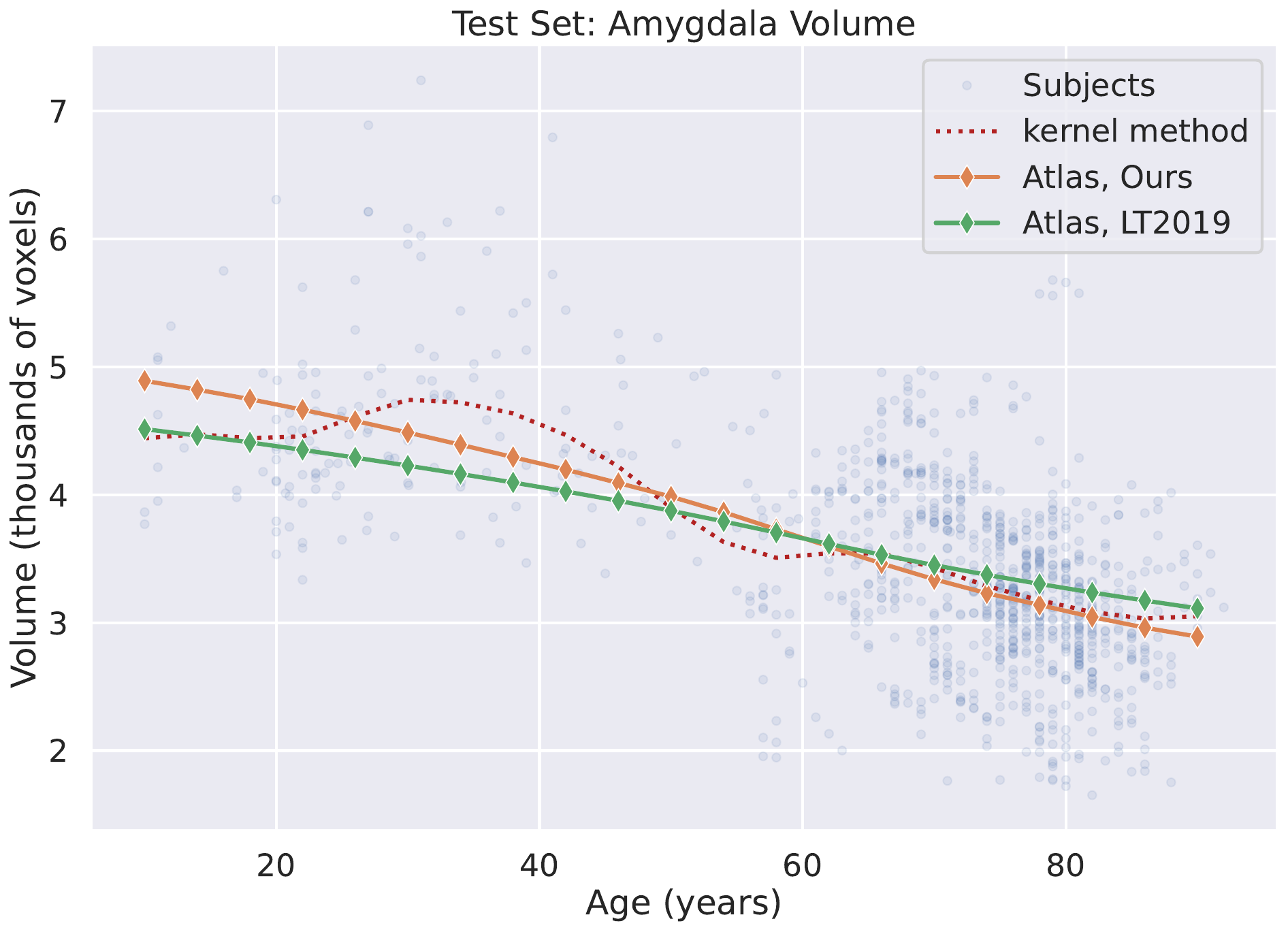}
    \includegraphics[width=0.23\textwidth]{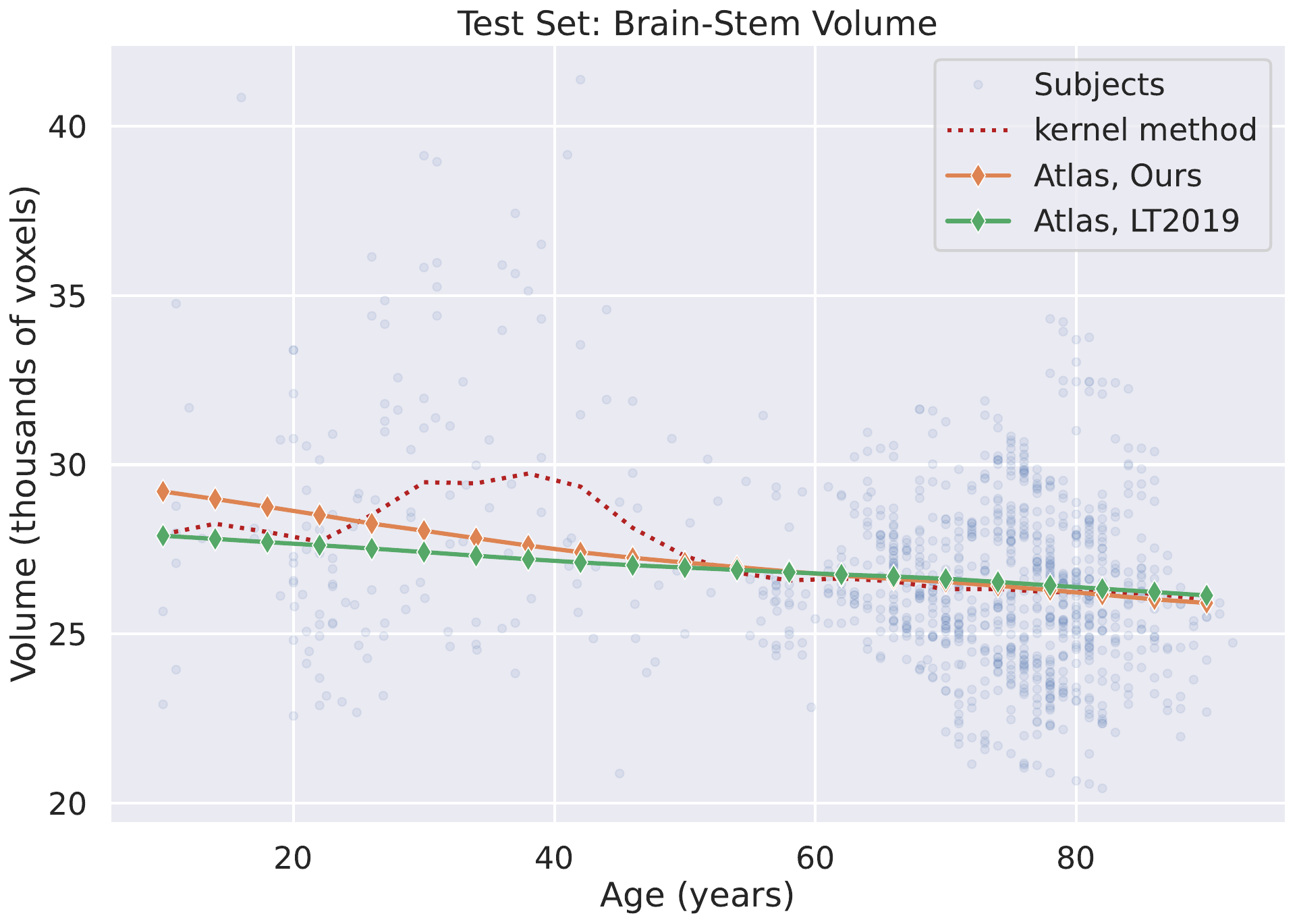}
    \includegraphics[width=0.23\textwidth]{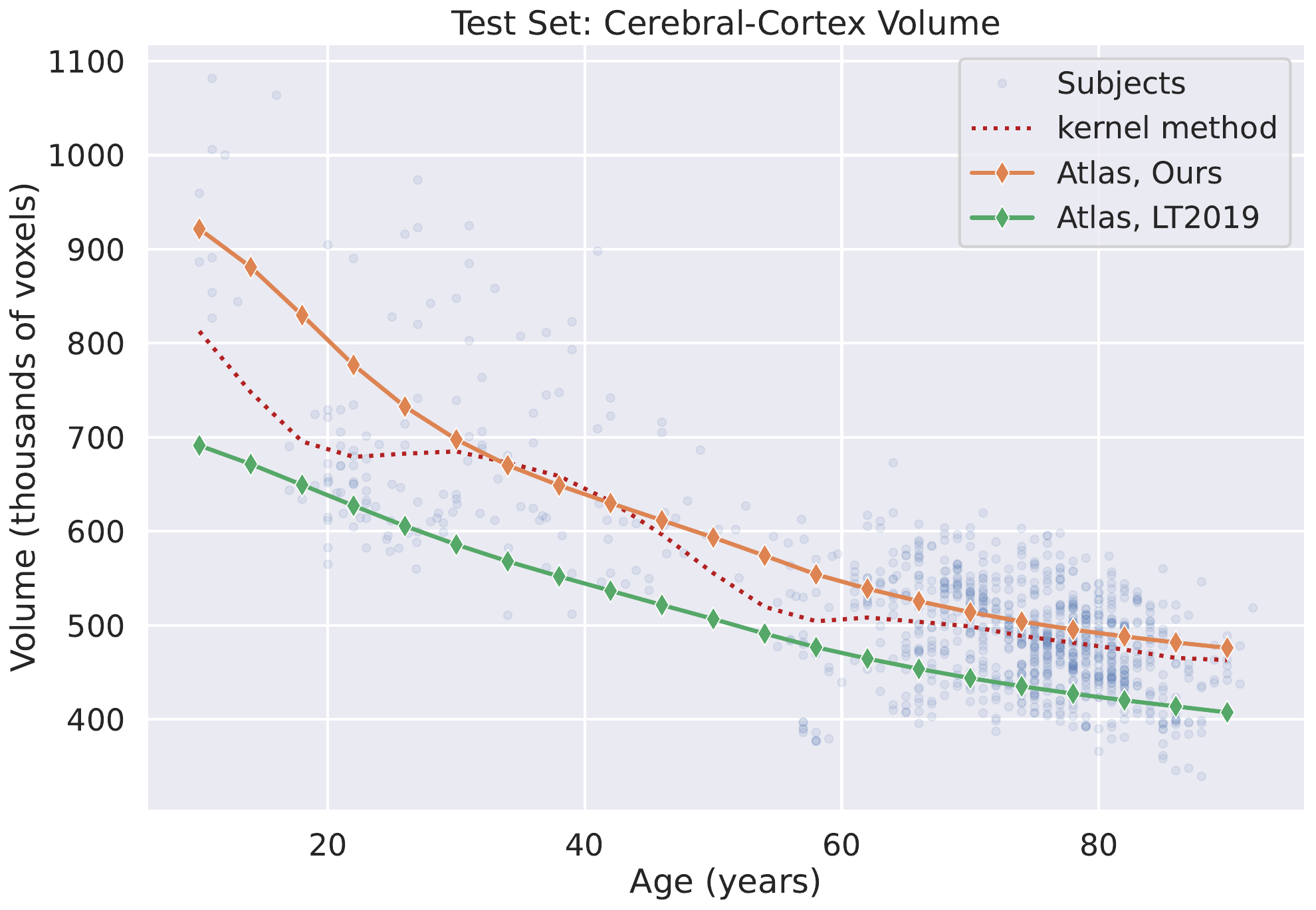}
    \includegraphics[width=0.23\textwidth]{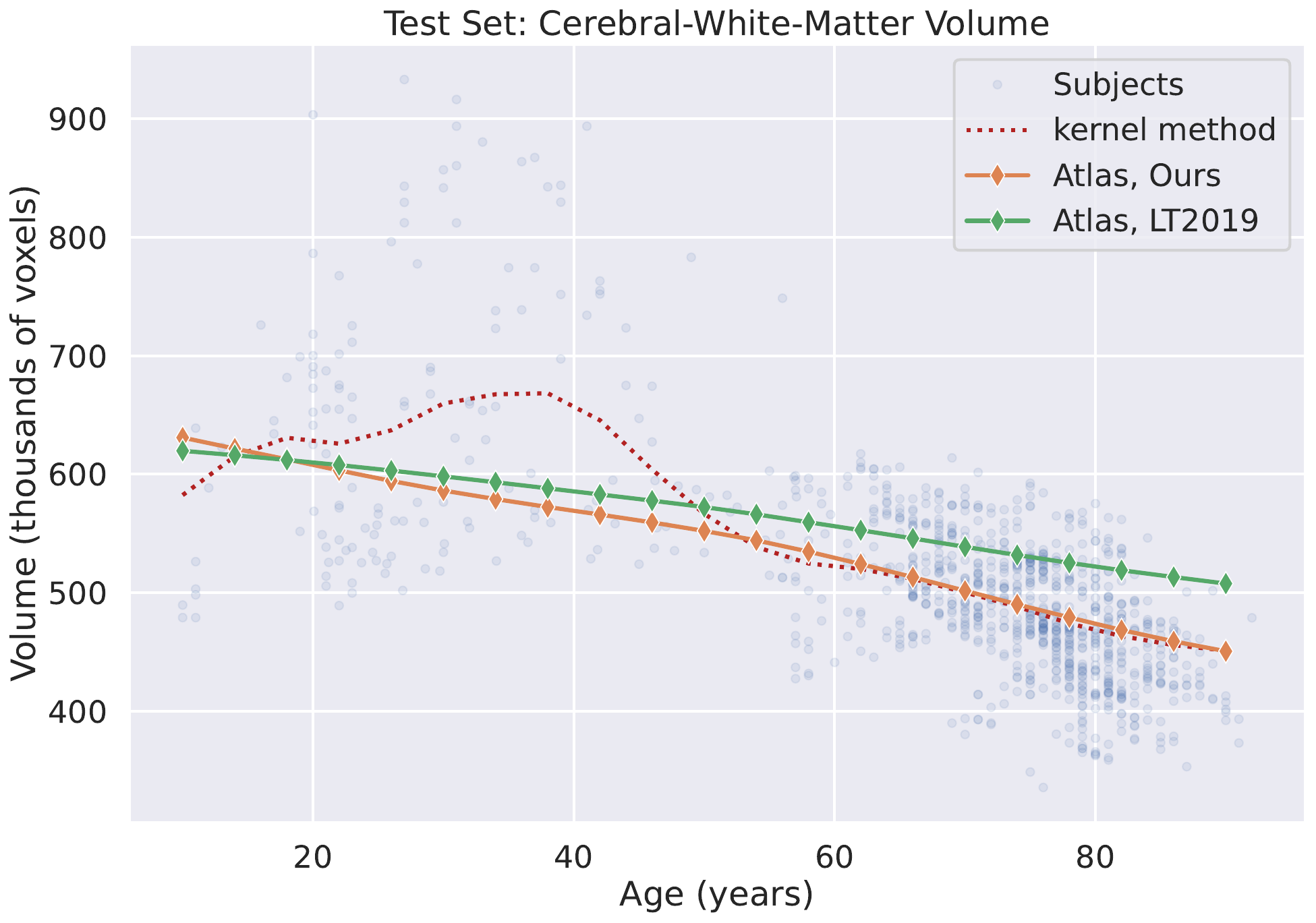}
    \includegraphics[width=0.23\textwidth]{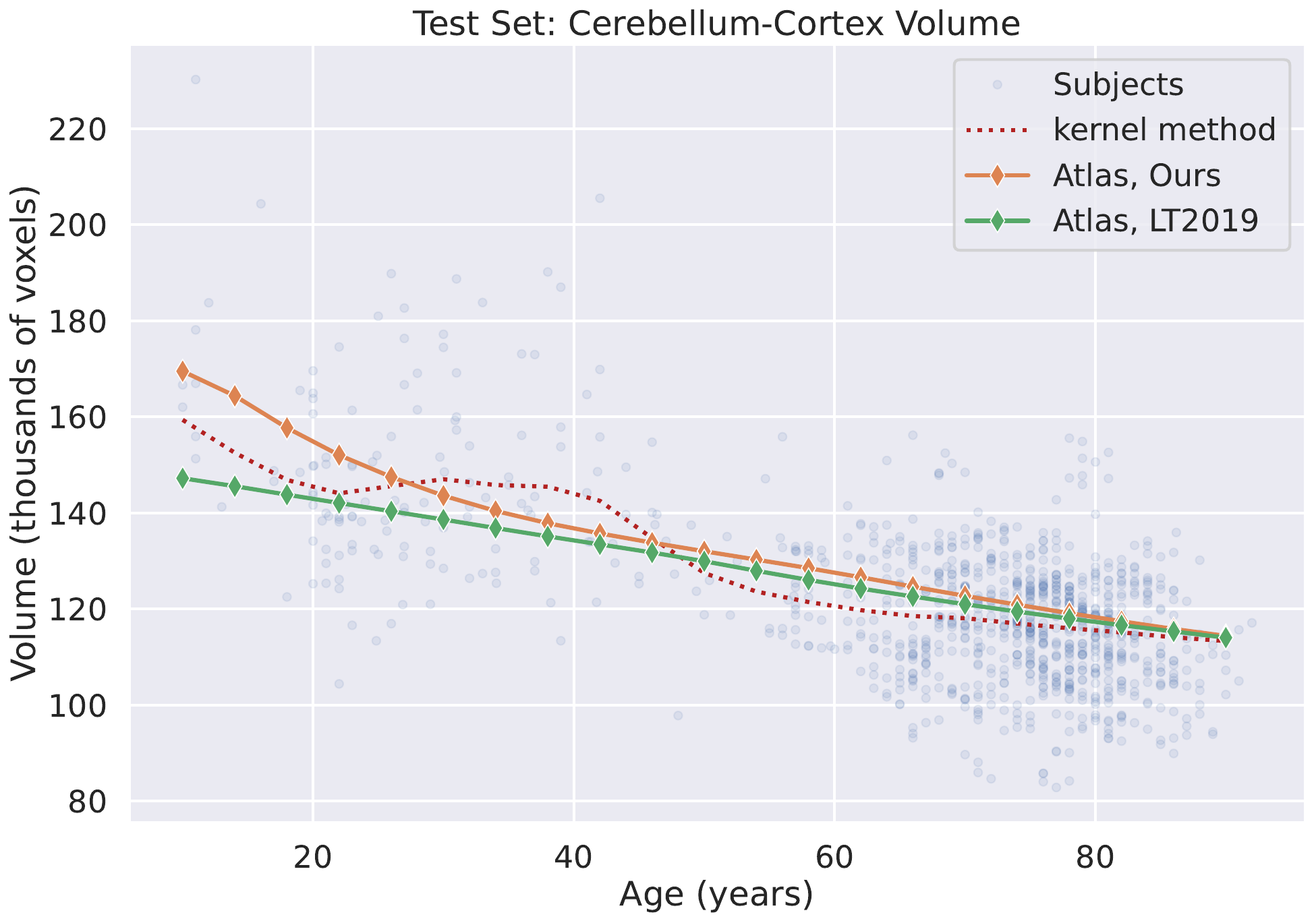}
    \includegraphics[width=0.23\textwidth]{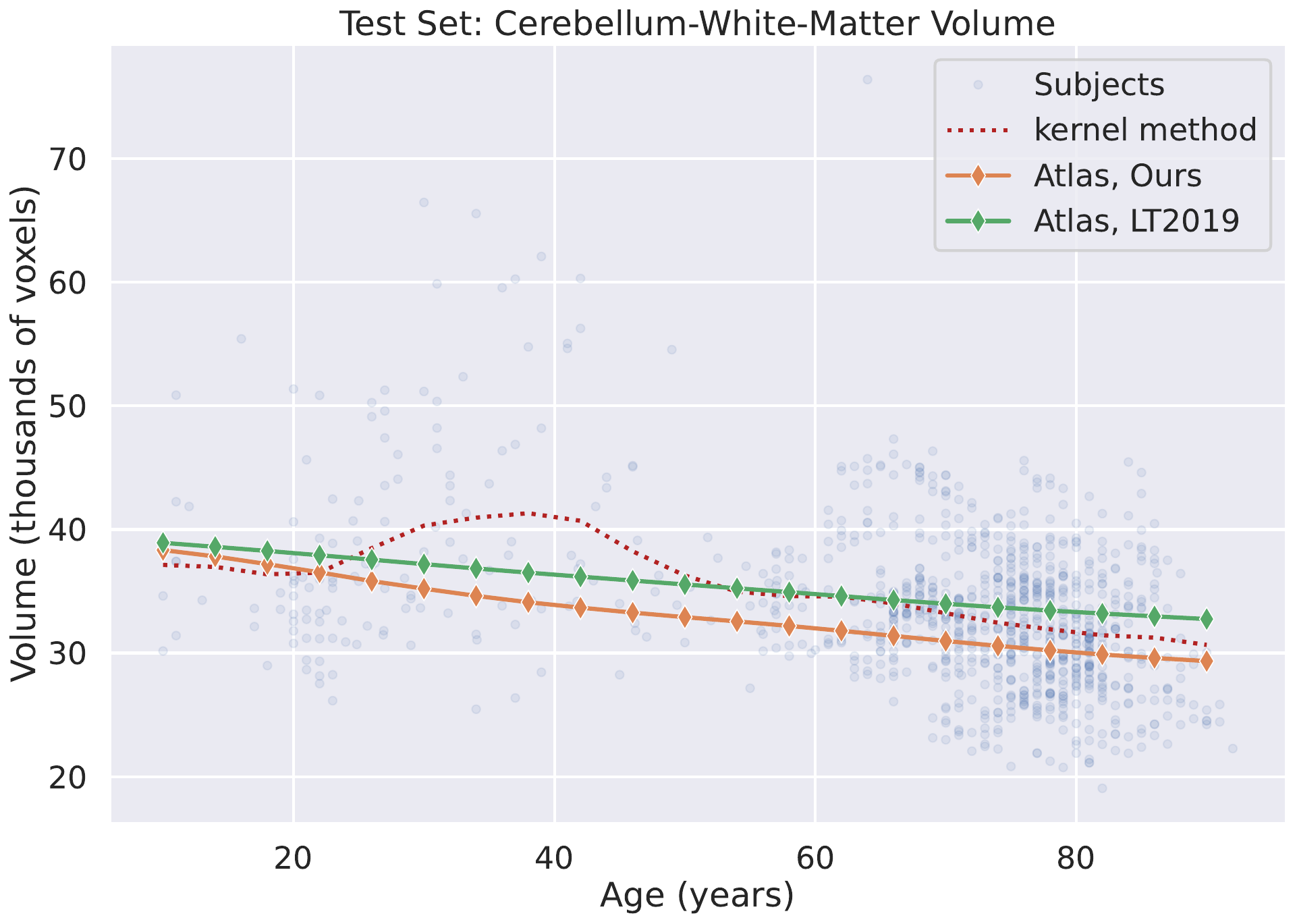}
    \includegraphics[width=0.23\textwidth]{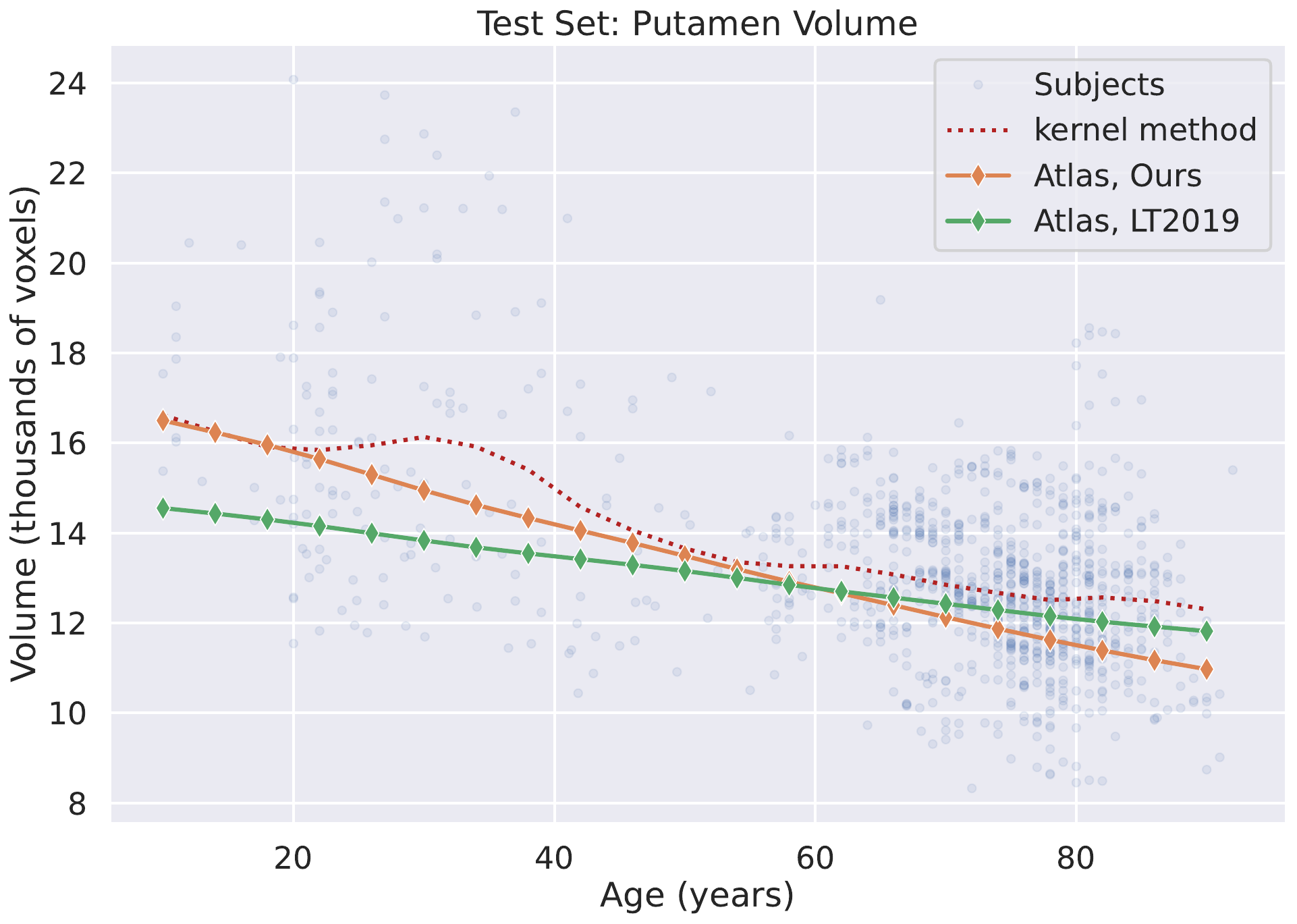}
    \includegraphics[width=0.23\textwidth]{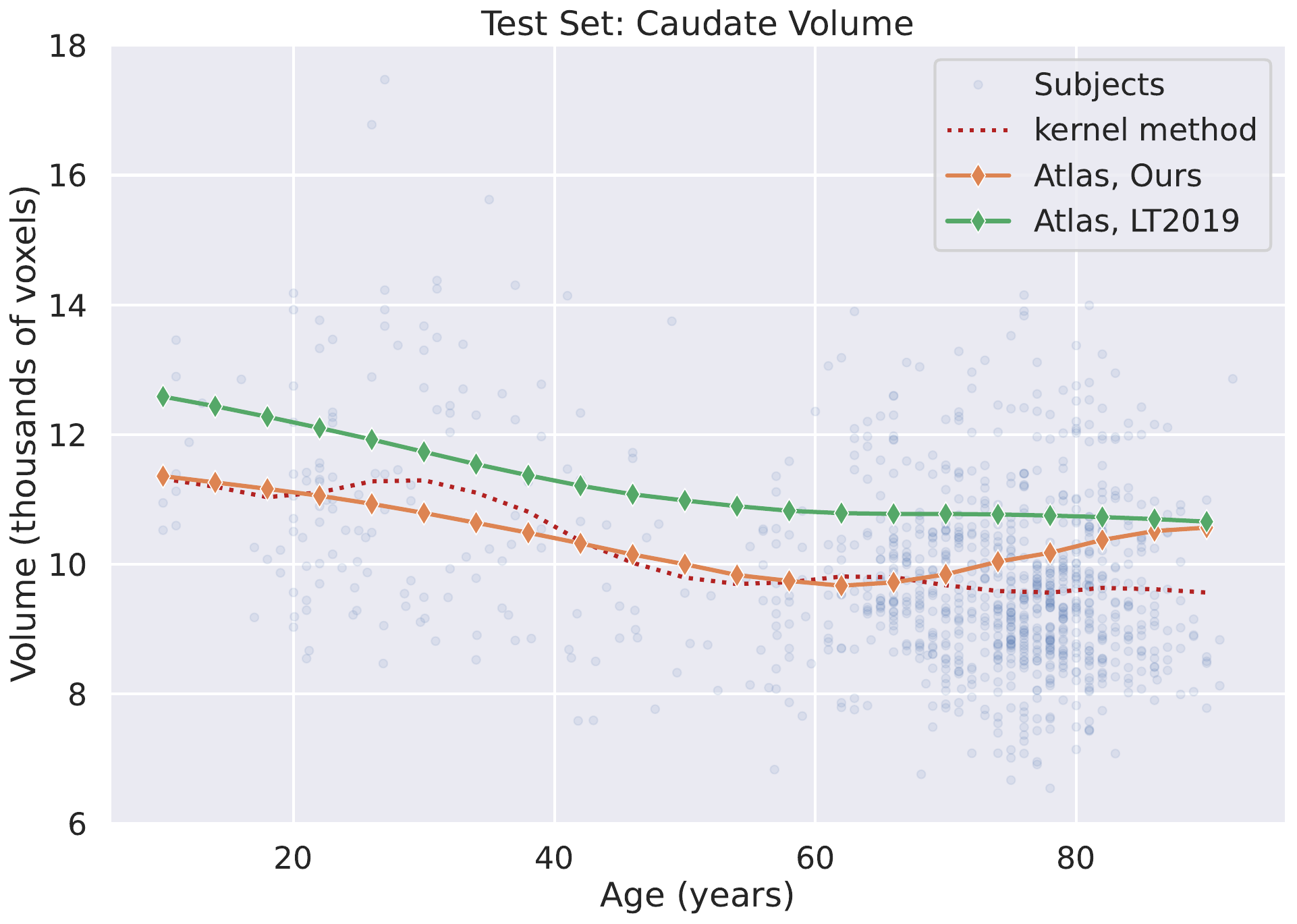}
    \includegraphics[width=0.23\textwidth]{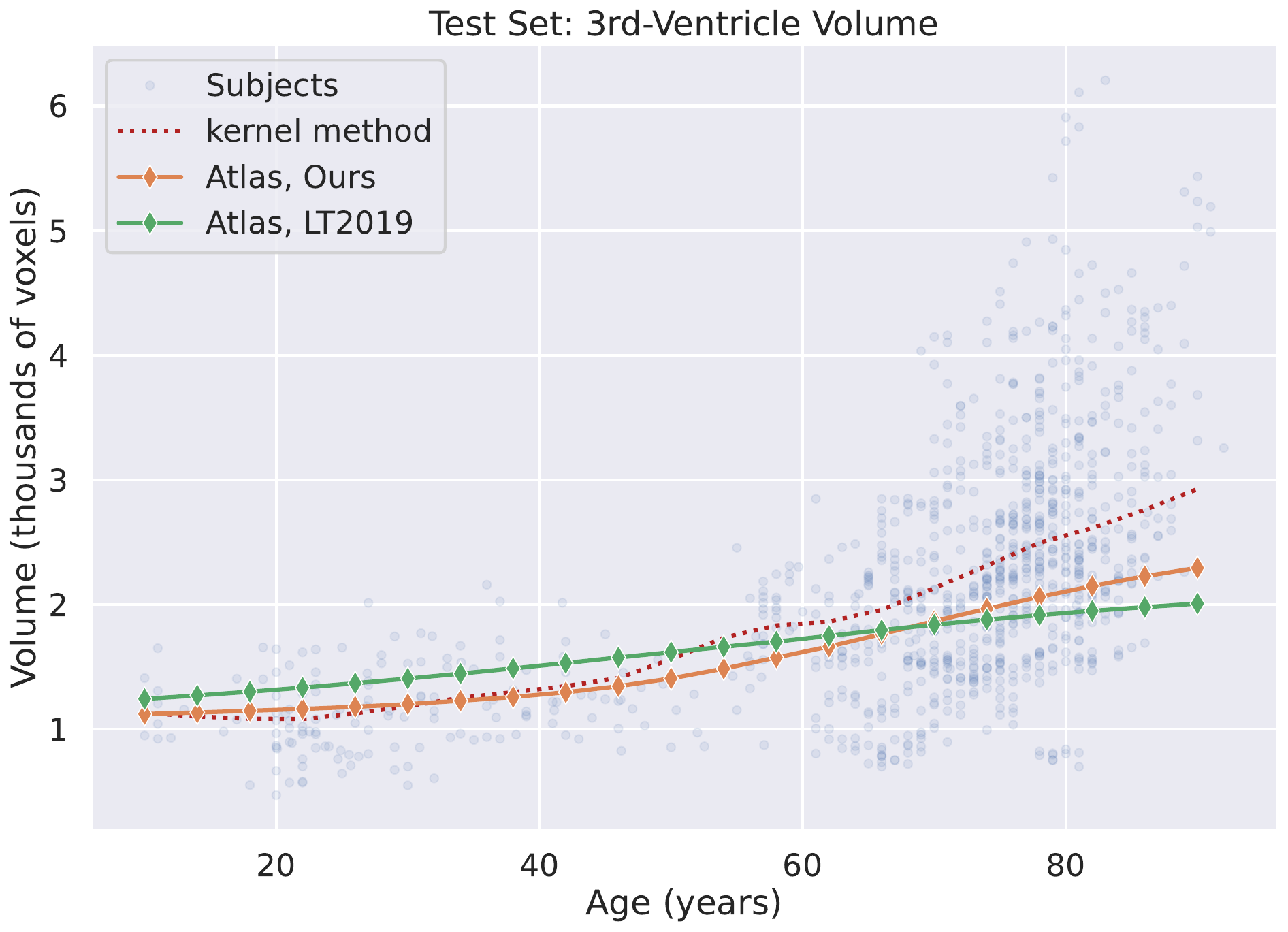}
    \includegraphics[width=0.23\textwidth]{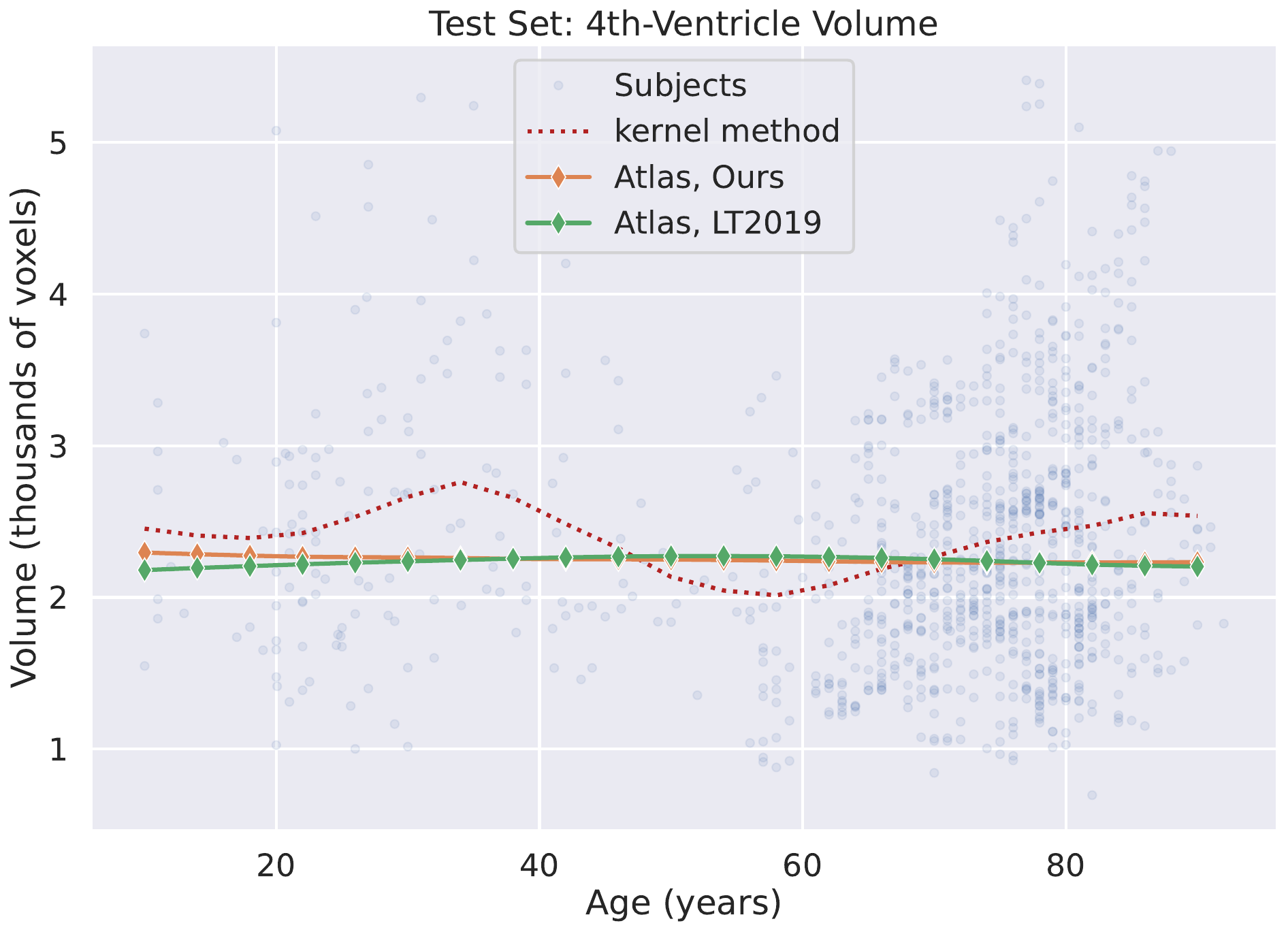}
    \includegraphics[width=0.23\textwidth]{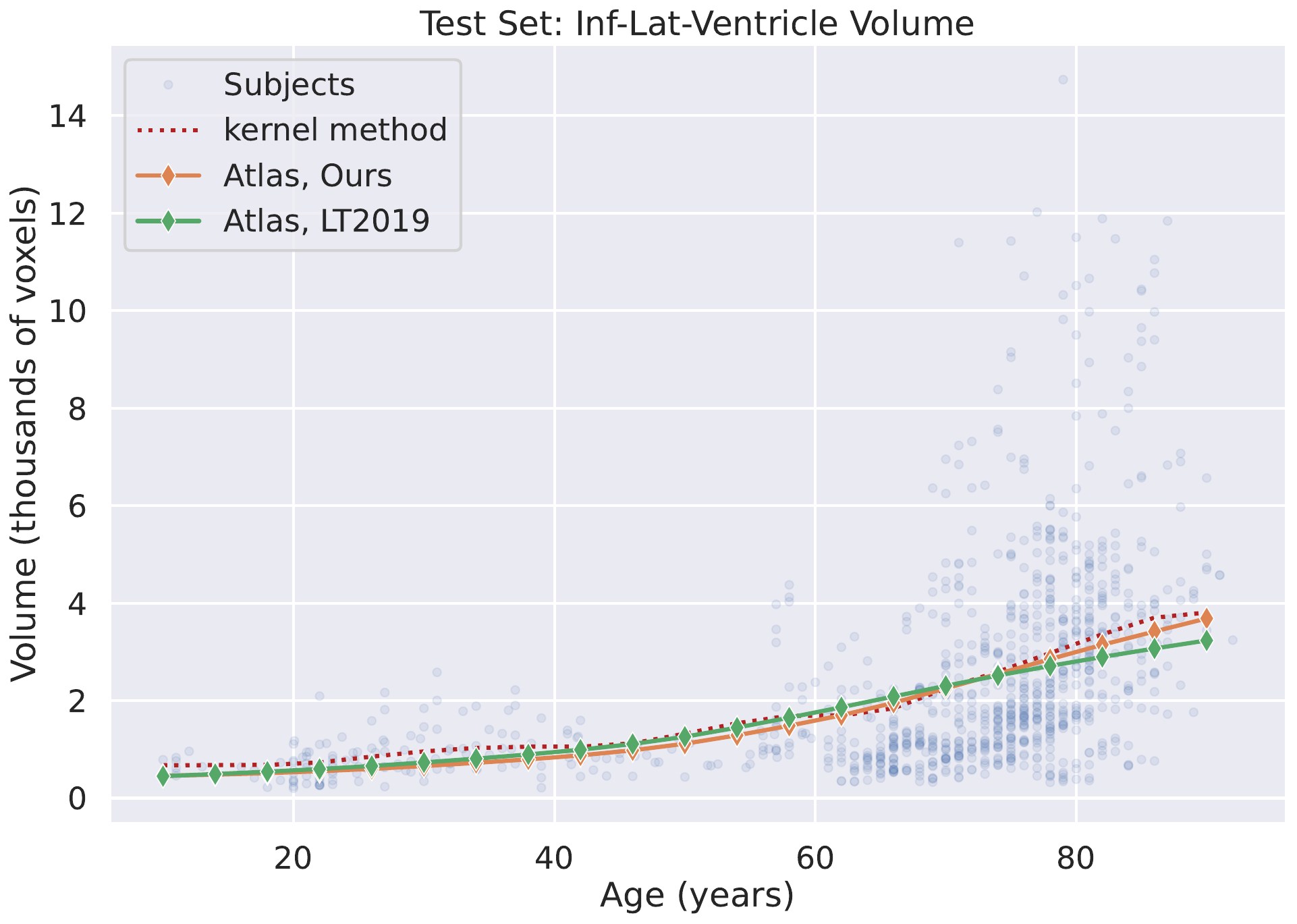}
    \includegraphics[width=0.23\textwidth]{final_figs/centrality_LT2019/M3D_Gaussian10_vol_all_Lateral-Ventricle_test_with_neuripsLT_no_mean.pdf}
    \includegraphics[width=0.23\textwidth]{final_figs/centrality_LT2019/M3D_Gaussian10_vol_all_Hippocampus_test_with_neuripsLT_no_mean.pdf}
    \includegraphics[width=0.23\textwidth]{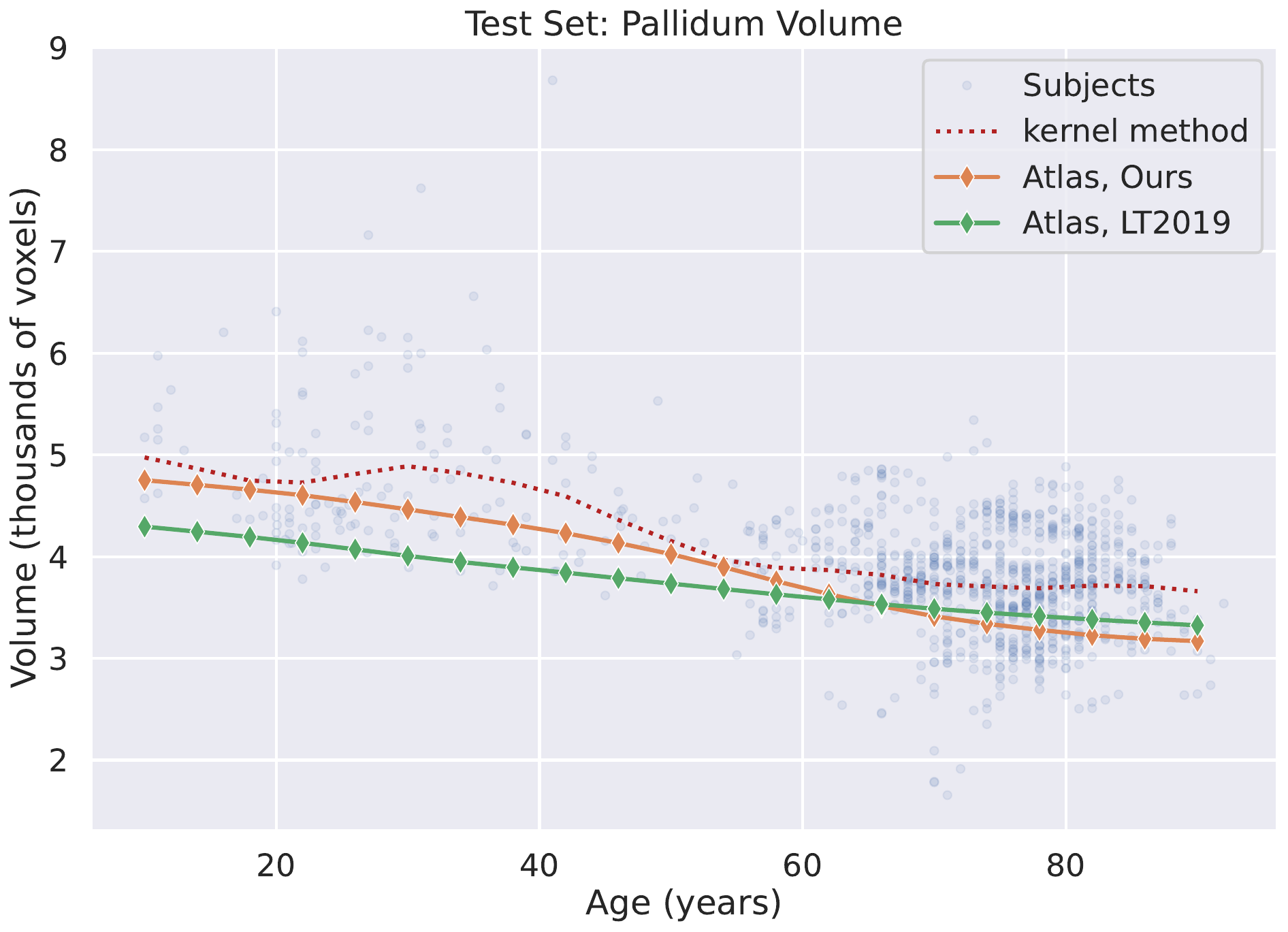}
    \includegraphics[width=0.23\textwidth]{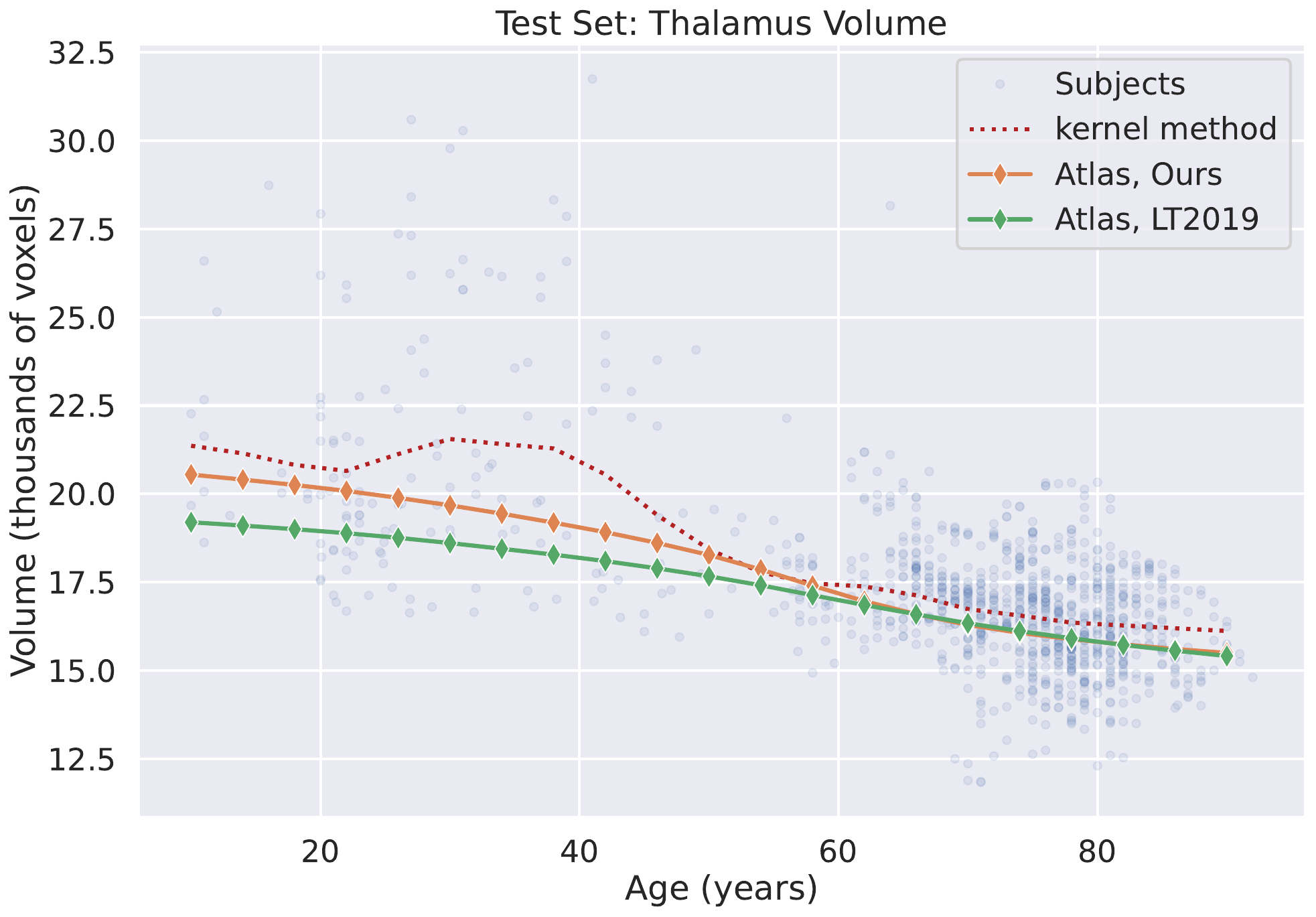}
    \includegraphics[width=0.23\textwidth]{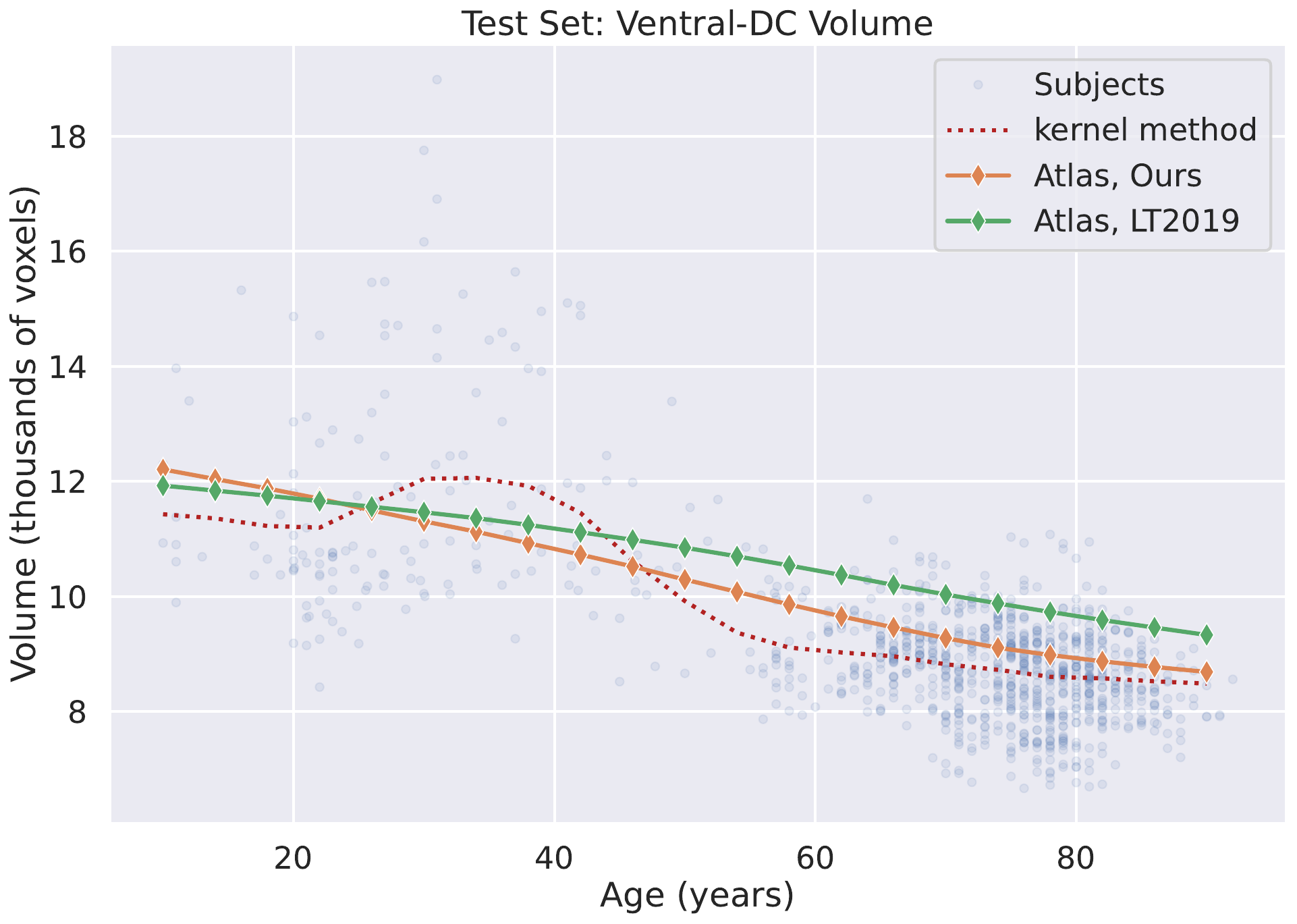}
    
    \caption{\textbf{Population trends for the brain structures in our data comparing \textit{AtlasMorph} and LT2019~\cite{dalca2019learning}.} The blue dots show the structure size for each brain MRI, sorted by the age of the subject when the MRI was taken. The starred orange line represents the volume captured by \textit{AtlasMorph}. The volume captured by a kernel method is shown in red and the one captured by LT2019~\cite{dalca2019learning} in green. \textit{AtlasMorph} captures more closely the population than LT2019.}
    \label{fig:sup_add_struct_pop_trend}
\end{figure*}

\begin{figure*}[h!]
  \begin{minipage}[t]{\textwidth}
    \centering
    \includegraphics[width=0.49\textwidth]{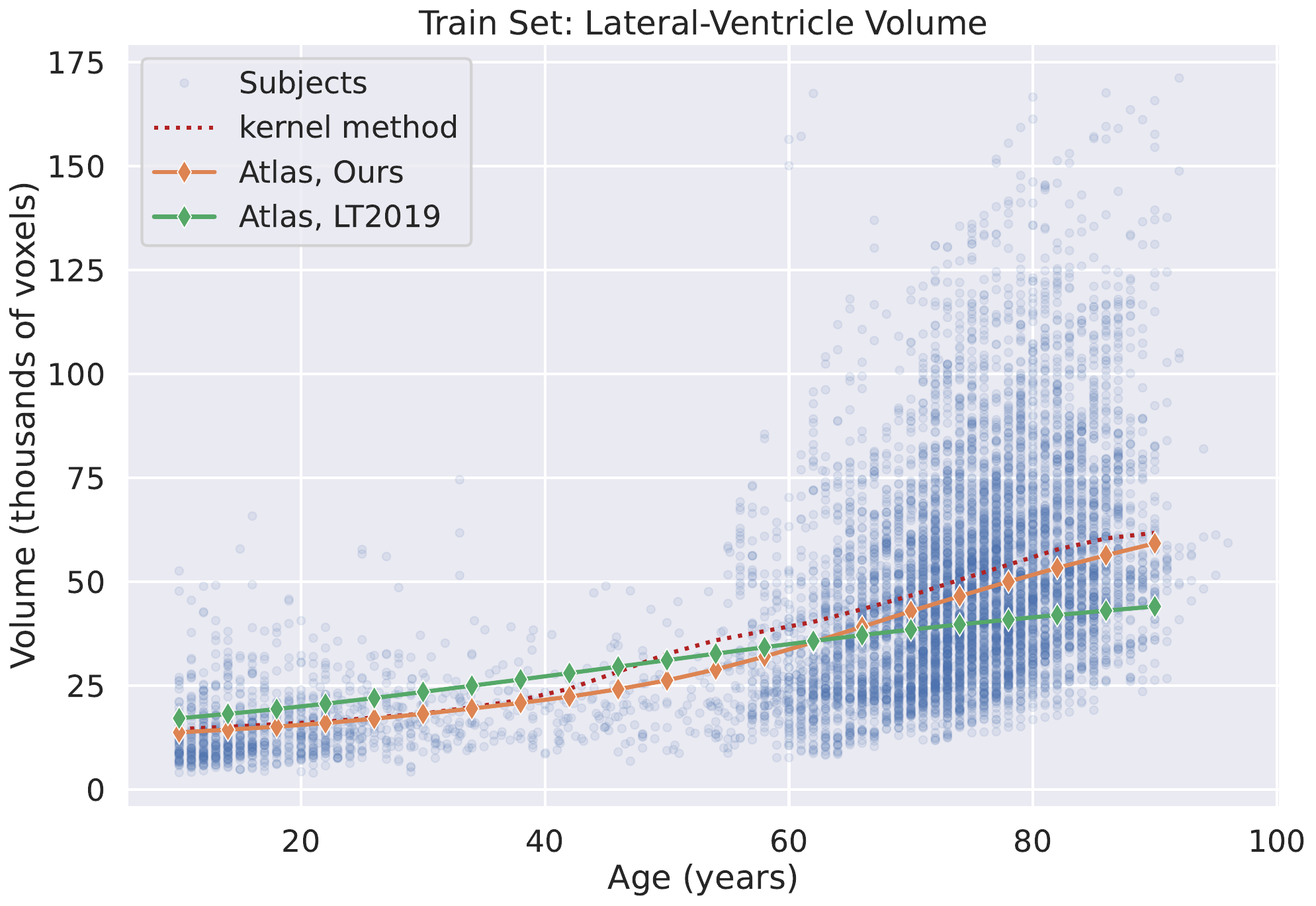}\hfill
    \includegraphics[width=0.49\textwidth]{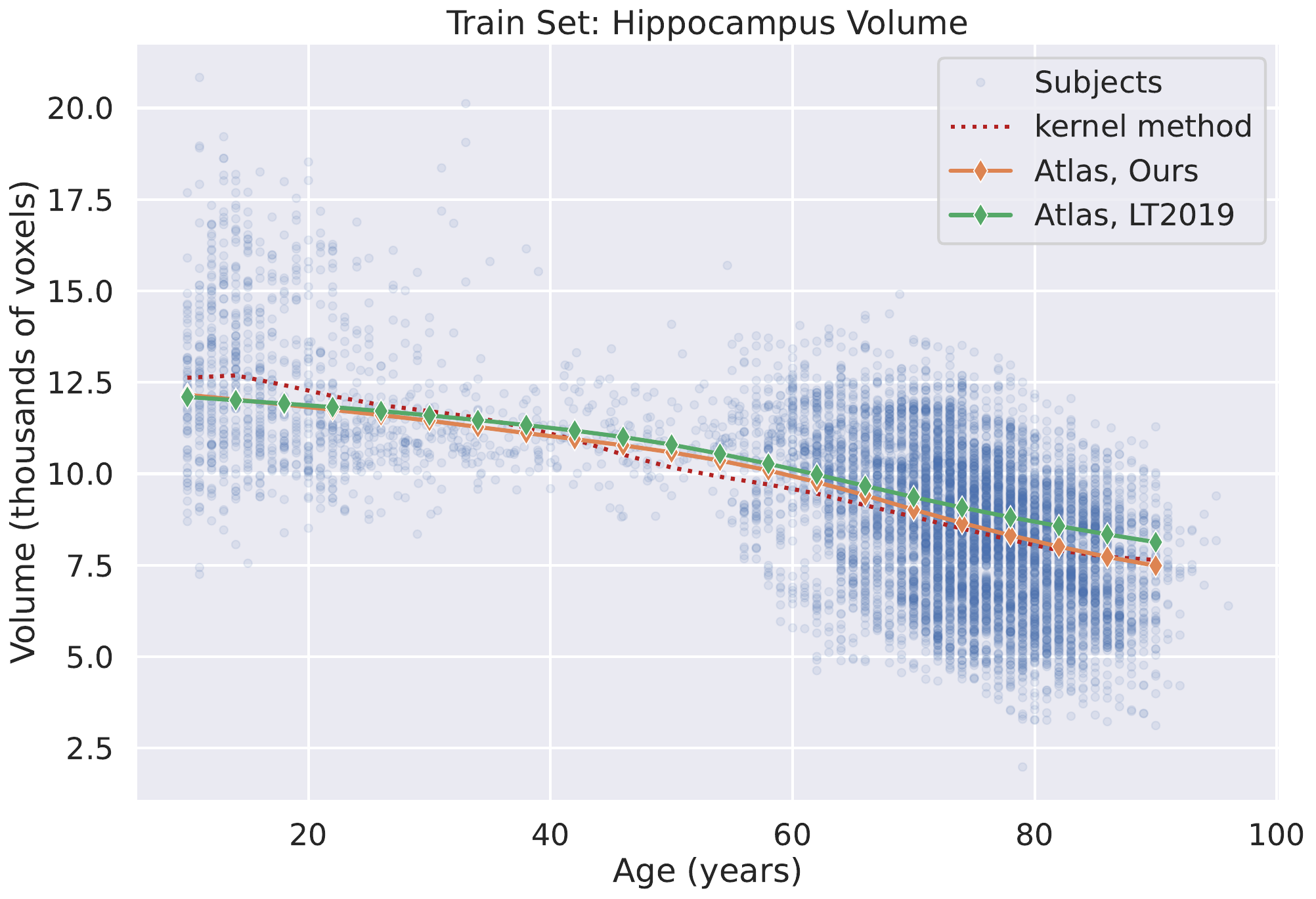}
    \caption{\textbf{Centrality using ventricles and hippocampi volumes conditioned on age for the training set.} The blue dots show the structure size (ventricles on the left, hippocampi on the right) for each brain MRI, sorted by the age of the subject when the MRI was taken. The starred orange line represents the volume captured by \textit{AtlasMorph}. The volume captured by a kernel method is shown in red and the one captured by LT2019~\cite{dalca2019learning} in green. \textit{AtlasMorph} captures more closely the population than LT2019.}
    \label{fig:3D_vent_vol_kernel_train}
  \end{minipage}
  \end{figure*}

\subsection{Results on the training set}
We include here additional analysis of our template using the training data, that contains significantly more subjects than the test data. In Figure \ref{fig:3D_vent_vol_kernel_train}, we show how \textit{AtlasMorph} can match population trends for both the ventricles and hippocampi structures. We also show that \textit{AtlasMorph} is more central that LT2019 in Figure \ref{fig:3D_vent_vol_kernel_train} and quantitatively in Figure \ref{fig:3D_strct_vol_train}, where we compute the relative error between the templates and the annotated labels.

\subsection{Population trends}
In Figure \ref{fig:sup_add_struct_pop_trend}, we present population trends capabilities of \textit{AtlasMorph} compared to the template proposed in LT2019~\cite{dalca2019learning} as a function of age for the following additional structures: amygdala, caudate, putamen, 3rd and 4th ventricles.

\end{document}